\documentclass[runningheads]{llncs}

\usepackage{eccv}

\usepackage{eccvabbrv}
\usepackage{adjustbox}
\usepackage{graphicx}
\usepackage{booktabs}
\usepackage{tabularx} 
\usepackage{rotating} 
\usepackage{svg}
\usepackage{longtable} 

\usepackage[accsupp]{axessibility}  

\usepackage{hyperref}

\usepackage{orcidlink}
\usepackage{booktabs} 
\usepackage{tabularx} 
\usepackage{pdflscape} 

\begin{document}


\title{Deciphering the Role of Representation Disentanglement: Investigating Compositional Generalization in CLIP Models}


\author{Reza Abbasi \and
Mohammad Hossein Rohban \and
Mahdieh Soleymani Baghshah}

\authorrunning{R. Abbasi et al.}

\institute{Sharif University of Technology, Tehran, Iran \\
\email{\{reza.abbasi,rohban,soleymani\}@sharif.edu}\\
\url{http://www.sharif.edu}}
\maketitle

\begin{abstract}
CLIP models have recently shown to exhibit Out of Distribution (OoD) generalization capabilities. However, Compositional Out of Distribution (C-OoD) generalization, which is a crucial aspect of a model's ability to understand unseen compositions of known concepts, is relatively unexplored for the CLIP models. Our goal is to address this problem and identify the factors that contribute to the C-OoD in CLIPs. We noted that previous studies regarding compositional understanding of CLIPs frequently fail to ensure that test samples are genuinely novel relative to the CLIP training data.
To this end, we carefully synthesized a large and diverse dataset in the single object setting, comprising attributes for objects that are highly unlikely to be encountered in the combined training datasets of various CLIP models.
This dataset enables an authentic evaluation of C-OoD generalization. Our observations reveal varying levels of C-OoD generalization across different CLIP models.
We propose that the disentanglement of CLIP representations serves as a critical indicator in this context. By utilizing our synthesized datasets and other existing datasets, we assess various disentanglement metrics of text and image representations. Our study reveals that the disentanglement of image and text representations, particularly with respect to their compositional elements, plays a crucial role in improving the generalization of CLIP models in out-of-distribution settings. This finding suggests promising opportunities for advancing out-of-distribution generalization in CLIPs. For more details and access to our dataset, please visit \href{https://github.com/abbasiReza/CLIP-COoD}{https://github.com/abbasiReza/CLIP-COoD}.

\end{abstract}

\section{Introduction}

\begin{figure}[t]
  \centering
  \includegraphics[width=0.9\textwidth]{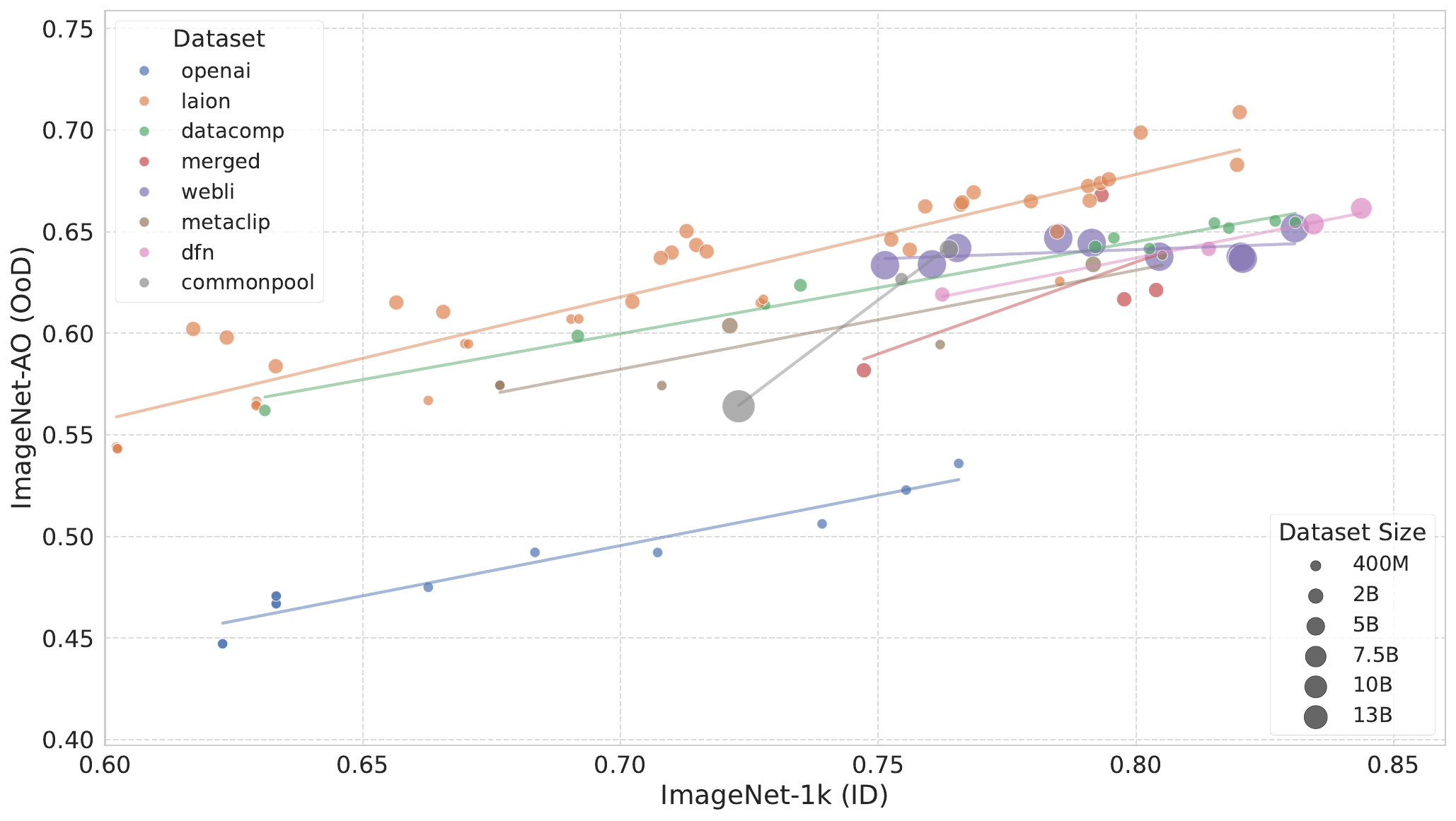}
 
  \caption{ Comparing zero-shot compositional out-of-distribution (C-OoD) generalization across diverse CLIP models and training sets. In-distribution (ID) performance is evaluated on the ImageNet validation set with object name labels, while the C-OoD generalization is assessed on our designed compositional dataset using attribute-object pair labels. Noticeably, CLIP models trained on the Common Pool dataset exhibit a steeper accuracy slope when transitioning from the ID to the OoD compositional setting compared to models trained on other datasets like WebLI. CLIPs trained on the LAION and DataComp datasets also show significantly higher C-OoD across ID accuracy.   Despite improved in-distribution accuracy, models pretrained on WebLI do not demonstrate substantial gains in generalizing to the novel compositional out-of-distribution test cases.}

  \label{fig:myfigure}
\end{figure}

Out-of-Distribution (OoD) generalization which is the ability of a model to generalize to the data distributions differing from the training distribution is very important for most learning models \cite{liu2021towards}.
In recent years, several studies suggested that some Vision-Language Models (VLMs) such as the CLIPs ~\cite{radford2021learning}, exhibit OoD generalization \cite{radford2021learning, fang2022data}. Specifically, several studies reported that CLIP models demonstrate enhanced zero- and few-shot accuracies on parallel versions of ImageNet, comprising images with various style shifts with respect to the original ImageNet \cite{fang2022data, nguyen2022quality}.

In particular, Compositional OoD (C-OoD) generalization is a main branch of the OoD generalization, focusing specifically on the ability of models to generalize to unseen combinations of known concepts or entities. Essentially, compositional generalization relates to human-like inductive biases that leads to more efficient learning via composing seen concepts \cite{wang-hershcovich-2023-evaluating}.
Recently, some studies have worked on evaluating or improving compositional generalization in the NLP tasks \cite{shaw-etal-2021-compositional, wang-hershcovich-2023-evaluating, mehta2021improving}. However, C-OoD generalization for vision tasks is less explored since the unseen compositions of concepts can not be easily created visually for investigation. 


In the recent years, evaluating the ability of VLMs in encoding objects, attributes, and their relations has recently received attention  \cite{yuksekgonul2023when, lewis2023does}.
Some benchmarks such as VL-Checklist \cite{zhao2023vlchecklist}, Winoground \cite{thrush2022winoground}, and Attribute-Relation-Order (ARO) \cite{yuksekgonul2023when} have been introduced to assess the image-text matching ability of VLMs in compositional setups more exactly. 
VL-Checklist provides a benchmark to evaluate VLMs capabilities in three categories of objects, attributes, and relations. ARO showcases that the reordering of words in the text does not highly impact on the similarity of the text with the corresponding image.
Some of these studies \cite{yuksekgonul2023when, thrush2022winoground} discussed shortcommings of VLMs
in encoding the compositional relationships between objects and attributes and \cite{lewis2023does} showed that VLMs
 can compose concepts
in a single-object setting including single attribute-object compositions.
Nonetheless, most of the work around compositional reasoning \cite{ossowski2024prompting,zhang2024countercurate,wang2024enhancing,doveh2023dense} were more concerned about compositional understanding of the inputs, and less attention has been paid to the OoD generalization in which the generalization ability are evaluated against truly novel compositions with respect to the training set.
In a nutshell, the literature suggests that compositional understanding in VLMs might be more feasible in the single-object setups. However, until now the C-OoD capability of CLIPs is unexplored. This makes us ask the question:

{\it Do CLIPs really have nontrivial C-OoD generalization in the single-object setting? and where does this ability stem from in such models?}

\begin{figure*}[t]
  \centering
  \includegraphics[width=\textwidth]{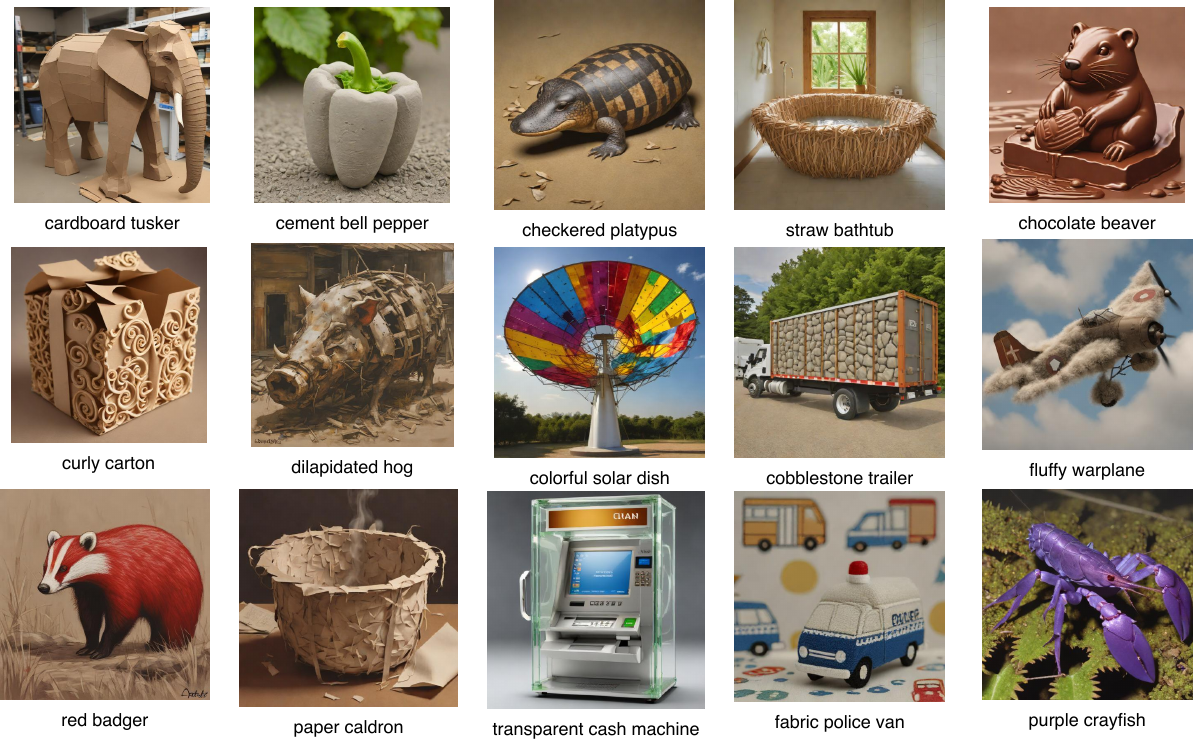} 

  \caption{Examples of images from our generated dataset. This dataset is created by combining attributes and objects that do not appear in the CLIP training sets, specifically designed for benchmarking compositional OoD generalization purposes.}
  \label{fig::dataset_design}
\end{figure*}

We propose a new benchmark to evaluate the C-OoD performance of CLIP models. Our approach involves generating a dataset, called ImageNet-AO (Attribute Object), distinct from the CLIPs training data. We gather comprehensive lists of objects and attributes, then generate images by combining these objects and attributes using a text-to-image model. The generated images undergo several filtering processes to ensure they are aligned with their intended and specified object-attribute description, and are novel compared to the combined CLIP training datasets both in the text and image domains.
We then evaluate different CLIP models on our OoD dataset to classify an input image into its composition constituents. Fig. \cref{fig:myfigure} gives an overview of this result, in which certain CLIPs, such as the ones trained on the LAION and DataComp, yielded strong C-OoD performance.

Finally, we analyze the factors that contribute to better performance in our benchmark. We found that the CLIPs that show higher C-OoD generalization typically exhibit strong disentangled text representations with respect to the composition constituents. We backed this observation by assessing numerous disentanglement metrics, and the intrinsic dimensionality of the  composition text embeddings.   We found that  CLIPs with strong C-OoD accuracy also enjoy a more disentangled image representation, albeit at a lower level compared to that of the text embedding. 
Based on these results, we hypothesize that the inherent disentanglement of the text is induced from the text representation space to that of the images through contrastive learning. We elaborate on this hypothesis in Sec. \ref{sec::CLIP_COOD}. 
Consistently,  various disentanglement metrics of the text and image representations are observed to be highly correlated in CLIPs. We also repeat all these experiments in datasets that were previously designed for evaluating disentanglement, and contain factors at a more fine-grained level, and note that all these observations hold.


Our contributions are summarized as follows:
\begin{itemize}
    \item Designing an image test dataset of attribute-object pairs that are unseen in common CLIP training datasets.
    \item Benchmarking the compositional generalization of various CLIPs in the carefully designed and controlled setting.
    \item Discovering that the CLIP representation space is decomposable into embedding of concepts (e.g., objects and attributes) especially for the embeddings obtained by the text encoder, and suggesting that it is the source of compositional generalization.
    \item Demonstrating a strong connection between CLIPs text/image disentanglements and better C-OoD generalization through different disentanglement metrics, on both our ImageNet-AO datasets and exisiting datasets designed previously for disentanglement evaluation.
\end{itemize}

\section{Methodology}
In this section, we explain how we conducted our study step-by-step. We first describe how we created our challenging benchmark dataset, ImageNet-AO, which involves finding new combinations and making images with text-to-image models (Sec. \ref{sec:design}). Examples of images in ImageNet-AO are shown in Fig. \ref{fig::dataset_design}. Then, we dive into how we test CLIP models in the zero-shot setting, and the chosen criteria to evaluate the models (Sec. \ref{sec:criteria}). 


\subsection{ImageNet-AO Dataset Design}
\label{sec:design}
\begin{figure*}[t]
  \centering
  \includegraphics[width=\textwidth]{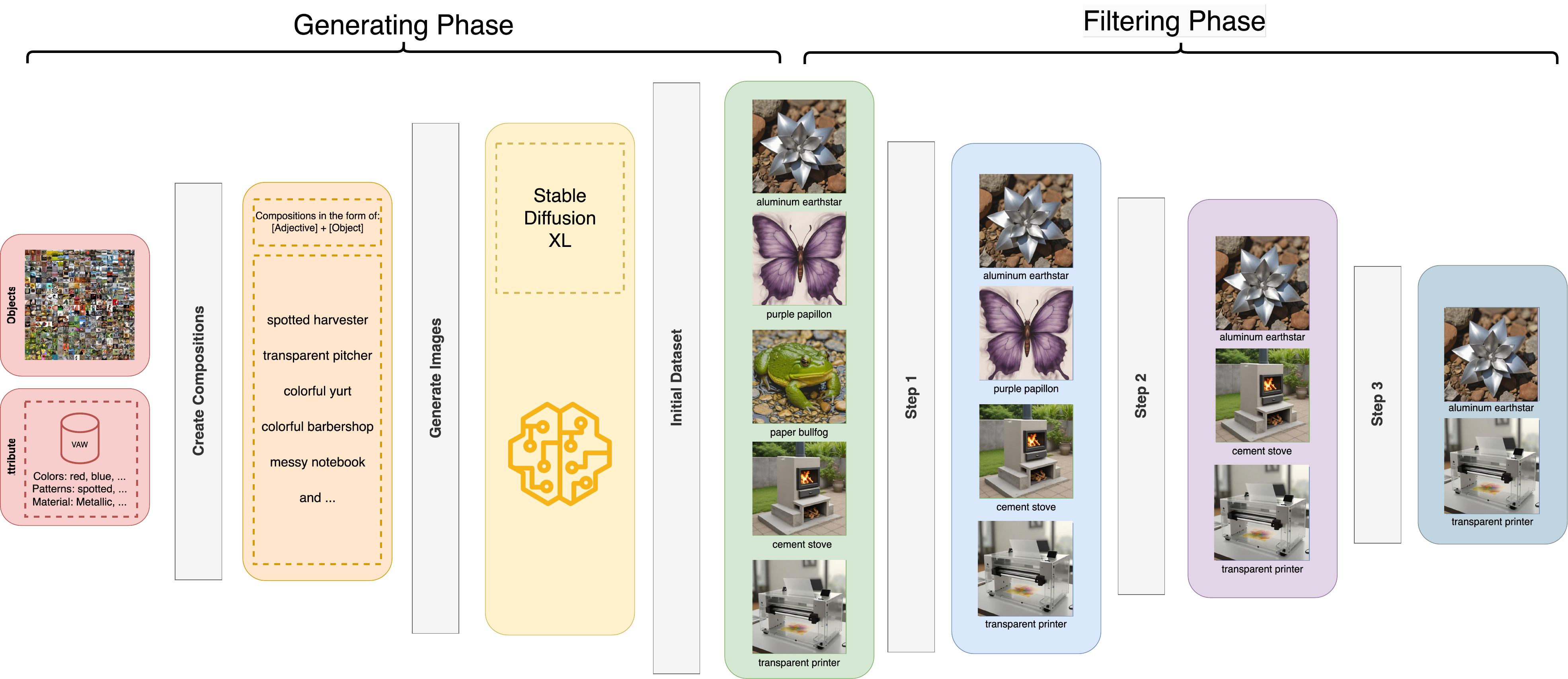}
  \caption{Dataset Design Stages: The data design process involves a generation phase  that makes the initial dataset from the whole set of the object and attribute compositions, and three distinct filtration steps. In the first filtration step, images where the target attribute or object lacks clear visibility are eliminated. In the second filtration step, the process removes images whose captions are already present in public datasets specifically curated for CLIP training. In the third filtration step, the faiss k-nearest neighbors algorithm is employed to identify and filter out images exhibiting similarities.}
  \label{fig:generate_and_filter}
\end{figure*}

To rigorously evaluate the compositional generalization capabilities of vision-language models, we devised an innovative dataset featuring compositions that are out-of-distribution  with respect to the training datasets of these models. Our dataset is crafted to include rare and unique compositions, thus ensuring it presents novel challenges to the VLMs under study. The dataset construction process is meticulously designed and involves several key steps, as depicted in \cref{fig:generate_and_filter} and detailed below:

\subsubsection{Selection of Objects (Nouns)}
Our initial step involved curating objects by extracting class names from the ImageNet dataset. This choice facilitates a direct comparison between the performance of models on our dataset and their performance on the well-established ImageNet validation set. By selecting a diverse array of class names, we aim to increase the complexity and richness of the generated compositional images.

\subsubsection{Selection of Attributes (Adjectives)}
We then selected 140 adjectives from the Visual Attributes Words (VAW) dataset\cite{pham2021learning}. These adjectives span various categories, including color, material, and texture, allowing us to create a wide range of descriptive combinations for image generation. A complete list of the 140 adjectives used from the Visual Attributes Words (VAW) dataset is provided in Appendix \ref{appendix_2}.

\subsubsection{Image Generation with Attribute-Object Prompts}
Utilizing the SD-XL Turbo, one of the most advanced and efficient text-to-image models available, we generated images based on combinations of the selected attributes and objects. By pairing 140 adjectives with 1,000 nouns, we created 140,000 unique prompts, which were then used to produce corresponding images, enriching our dataset with a vast array of compositional variety.





\subsubsection{Filtering Process}

To guarantee the integrity and the intended OoD characteristics of our dataset, we implemented a meticulous three-step filtering process. This approach ensures that our dataset not only accurately represents the specified attribute-object combinations but also stands apart from existing datasets in terms of composition and novelty. The steps are as follows:

\textbf{Step 1 - Initial Validation}:
Each generated image was subjected to an initial evaluation to verify its accuracy in depicting the intended attribute-object pair, exclusively through human assessment. During this process, evaluators were tasked with answering two critical questions: "Is this an image of [object]?" and "Does it exhibit [attribute]?" If at least one of these questions was answered with a "no," the image was removed from consideration. This step ensured that only images accurately representing the specified characteristics were retained for further processing.

\textbf{Step 2 - Exclusion of Known Combinations}:
To ensure the exclusivity of our dataset, we conducted a comprehensive search across several datasets (LAION, CommonPool, YFCC, and CC) to identify and eliminate any attribute-object combinations already present. This was achieved through a relaxed matching criterion, where combinations were removed if both the object and attribute appeared in a caption of an image, even if not in direct association.

\textbf{Step 3 - Verification of OoD Status}:
The final step in our filtering process was to ensure the OoD nature of our dataset. We used the Faiss library \cite{douze2024faiss} for a K-nearest neighbors search to compare our generated images against those in the LAION, CommonPool, YFCC, and CC datasets. Images were considered unique and retained in our dataset if no closely matching analogs were found, based on human evaluation. This rigorous approach ensured the novelty and uniqueness of our dataset by excluding combinations that had similar matches in the referenced datasets.

The dataset design process culminates in around 23,000 novel combinations of attributes and objects. The final generated dataset, after passing through the filtering process, comprises approximately 60,000 images representing  23,000 unique attribute-object combinations. Detailed properties and statistics about the dataset, including the list of attributes and objects used, can be found in the appendix.

\subsection{Model/Data Zoo and Evaluation Criteria} 
\label{sec:criteria}
In our experiments, we evaluate CLIP models trained on a diverse selection of datasets, including OpenAI's private dataset, LAION, YFCC15m, CC12m, DataComp, DFN-5B, WebLI, and CommonCrawl. These models leverage a variety of backbone image encoders such as ResNet50, ResNet101, ViT-B-32, ViT-B-16, ViT-L-14, ViT-H-14, ViT-g-14, and ViT-BigG-14. Our evaluation also extends to new CLIP variations, including EVA CLIP, SigLIP, and CLIPA, allowing for a comprehensive assessment of their performance and generalization capabilities across different tasks and datasets.


\section{Comparison of CLIP Models on ImageNet-AO} \label{sec:comparison_setup}
To evaluate the CLIP model performance in the 
 classification tasks, we adopted the evaluation method developed by \cite{ilharco_gabriel_2021_5143773}, similar to the zero-shot evaluation approach described in  \cite{radford2021learning}. Our evaluation involves providing the model with the actual images and various captions, obtaining embeddings for both the images and texts, and calculating their cosine similarities. This allows us to estimate the relevance of the captions to the image content, similar to a classification task. Given that our dataset only provided class labels (attribute-object pairs) for images, we expanded on this by creating 80 captions per class using various templates. This approach, inspired by the methodology described in \cite{radford2021learning}, allows for a more comprehensive representation of each class. We generated embeddings for these captions and averaged them to produce a final embedding for each class, which was then used in our zero-shot evaluation. For the test sets, all 1000 classes of ImageNet were used as the in-distribution set and expanded the number of classes to approximately 21000 for the OoD set. The CLIP evaluations are shown in Fig. \ref{fig:myfigure}.

While our results generally showed that models trained on larger datasets exhibited improved accuracy in both in-distribution and out-of-distribution settings, supporting the notion that larger training datasets can enhance compositional out-of-distribution generalization performance, it is crucial to note that dataset size alone does not directly predict model strength. The performance of models varied significantly with not only the dataset size but also the quality and curation of the data.
For instance, CLIP trained on the unfiltered CommonPool-XL dataset performed weaker than CLIP trained on the CommonPool-XL dataset filtered using ClipScore, despite the unfiltered dataset containing an additional 7 billion images. This further reinforces that simply increasing dataset size does not necessarily lead to improved model performance, and carefully curating and filtering the data can be more effective than merely accumulating vast amounts of unfiltered data.

Additionally, as evident from Fig. \ref{fig:myfigure}, models with different configurations trained on various datasets exhibited different training slope trajectories. The models trained on CommonPool-XL with different data filtering techniques demonstrated particularly steep performance trends, suggesting that the combination of a large dataset and effective data curation can yield significant performance gains.

Interestingly, the SigLip (denoted as WebLI) models presented a unique case with a somewhat negative slope, indicating that while enhancements to the backbone architecture improve in-distribution data performance, they may adversely affect out-of-distribution data performance. This highlights the nuanced relationship between architectural improvements and model generalization capabilities.




This extensive analysis, which encompasses the performance of diverse CLIP models across a broad spectrum of datasets, underscores the complexity of factors influencing model behavior and the pivotal role of dataset characteristics in achieving optimal performance in both in- and out-of-distribution settings. Further details on the performance evaluation of various CLIP models can be found in Sec. \ref{appendix_1} of the Appendix.

\section{Why CLIP has Compositional Generalization?} \label{sec::CLIP_COOD}

Having established the superior C-OoD performance of certain CLIPs, we next try to investigate the reasons behind these observations.  It has been widely known 
that disentangled representations make meaningful construction  of known concept mixtures in the embedding space feasible, hence resulting in better C-OoD generalization \cite{yang2023vectorbased,montero2021the, xu2022compositional}. Here, disentanglement means assignment of separate and independent embedding dimensions to different factors of variations, which in this case are the objects and attributes.

We hypothesize that the discrete nature of the language, and large and diverse training datasets promote a more decomposable text representation. On the other hand,  alignment of the text and image embeddings through contrastive learning in CLIPs induces this decomposability in the image domain.    
Based on these insights, we posit that representation decomposability is the key to the CLIP unseen compositional generalization. This claim is supported by two main arguments:
\begin{itemize}
\item Decomposability of the CLIP text embedding, measured through a comprehensive set of metrics, is correlated to the CLIP C-OoD generalization (Fig. \ref{fig:ood_vs_disentanglement}, bottom row).
\item Text representation disentanglement is induced in the image encoding, due to implicit maximization of the mutual information of text and image representations through contrastive learning. We elaborate on this claim empirically (Fig. \ref{fig:ood_vs_disentanglement}, top row), and theoretically in what follows. 
\end{itemize}

\subsubsection{Why disentanglement is induced from one view to another in the contrastive learning?} \label{sec::theory_dise}

We next try to give some theoretical insight on why and how the disentanglement emerges in the CLIP vision encoder. 
Several studies have shown the relation between minimizing the contrastive loss and maximizing the mutual information \cite{chen2020simple}. Therefore, the CLIP training implicitly maximizes the mutual information between text and image embeddings. We claim that disentanglement in the text representation, which was evidenced previously, may encourage disentanglement in the image encoding. To see this, let $y_1$ and $y_2$ be the text embeddings for the objects and attributes, respectively. Let $x_1$ and $x_2$ be the corresponding image embeddings. Assuming a decomposable text embedding means $y_1 \perp y_2$, i.e. $p(y_1,y_2)=p(y_1)p(y_2)$. Now by minimizing the contrastive loss, the mutual information $I(x_1,x_2;y_1,y_2)$ is maximized. By letting $x=(x_1,x_2)$, and $y=(y_1,y_2)$, we have:
\begin{multline*}
I(x_1, x_2; y_1, y_2) 
= \text{D}_{\text{KL}}(p(x, y) \parallel p(x) p(y)) \\
=  \text{D}_{\text{KL}}(p(x_1|x_2, y) p(x_2|y) p(y) \parallel p(x_1|x_2) p(x_2) p(y)) \\
=  \mathbb{E}_{x_1,x_2, y}[\log(p(x_1|x_2, y)/p(x_1|x_2))] + \mathbb{E}_{x_2, y}[\log(p(x_2|y)/p(x_2))]
\\
= \mathbb{E}_{x_2, y}[\text{D}_{\text{KL}}(p(x_1 | x_2, y) \parallel p(x_1 | x_2))] + \mathbb{E}_{y} [\text{D}_{\text{KL}}(p(x_2 | y) \parallel p(x_2))]
\end{multline*}

Maximization of the latter term makes $x_2$ and $y$ dependent random variables, otherwise if $x_2 \perp y$, the expected KL divergence would be minimum (or zero), which is against maximizing the mutual information. 
Note that however, $x_2$ does not ideally depend on both $y_1$ and $y_2$, otherwise the two distributions in the KL divergence in the first term become similar, which is also against maximizing the mutual information. Putting these together, $x_2$ mostly depends on $y_2$ if the mutual information is maximized. Using a symmetric argument, $x_1$ mostly depends on $y_1$. Finally, because $y_1 \perp y_2$, we conclude that $x_1$ and $x_2$ tend to become independent. Therefore, maximizing $I(x_1,x_2;y_1,y_2)$ decomposes $x$ if $y$ is already decomposed.

\section{Decomposable representation of CLIP Models} \label{sec::embd_disen}


In this section, our primary objective is to leverage the generated dataset and other synthtic datasets to analyze our hypotheses, focusing on the decomposable CLIP representation space and its impact on the compositional OoD performance.


\subsection{Attribute-Object Decomposition of Representation Space}
In this section, we show that the representation space of the CLIP models on the proposed dataset can be decomposable into the representations of the objects and the attributes.
\subsubsection{Disentanglement of Attributes and Objects}
Here, we aim to assess the level of embeddings disentanglement in various CLIPs on ImageNet-AO. We utilize some common disentanglement metrics, namely the Z-Diff Score \cite{higgins2016beta}, DCI \cite{eastwood2018framework} and  Explicitness score \cite{ridgeway2018learning} to quantitatively evaluate the embeddings. These metrics are typically employed for supervised disentanglement assessment and require access to the latent factors of data. Since we have a compositional text specifying the attribute and the object for each image, we can consider two super latent factors corresponding to attributes and objects respectively. More details about these disentanglement metrics and their formulas can be found in Appendix \ref{appendix_3}.

\begin{figure}[!t]
  \centering
  \includegraphics[width=\textwidth]{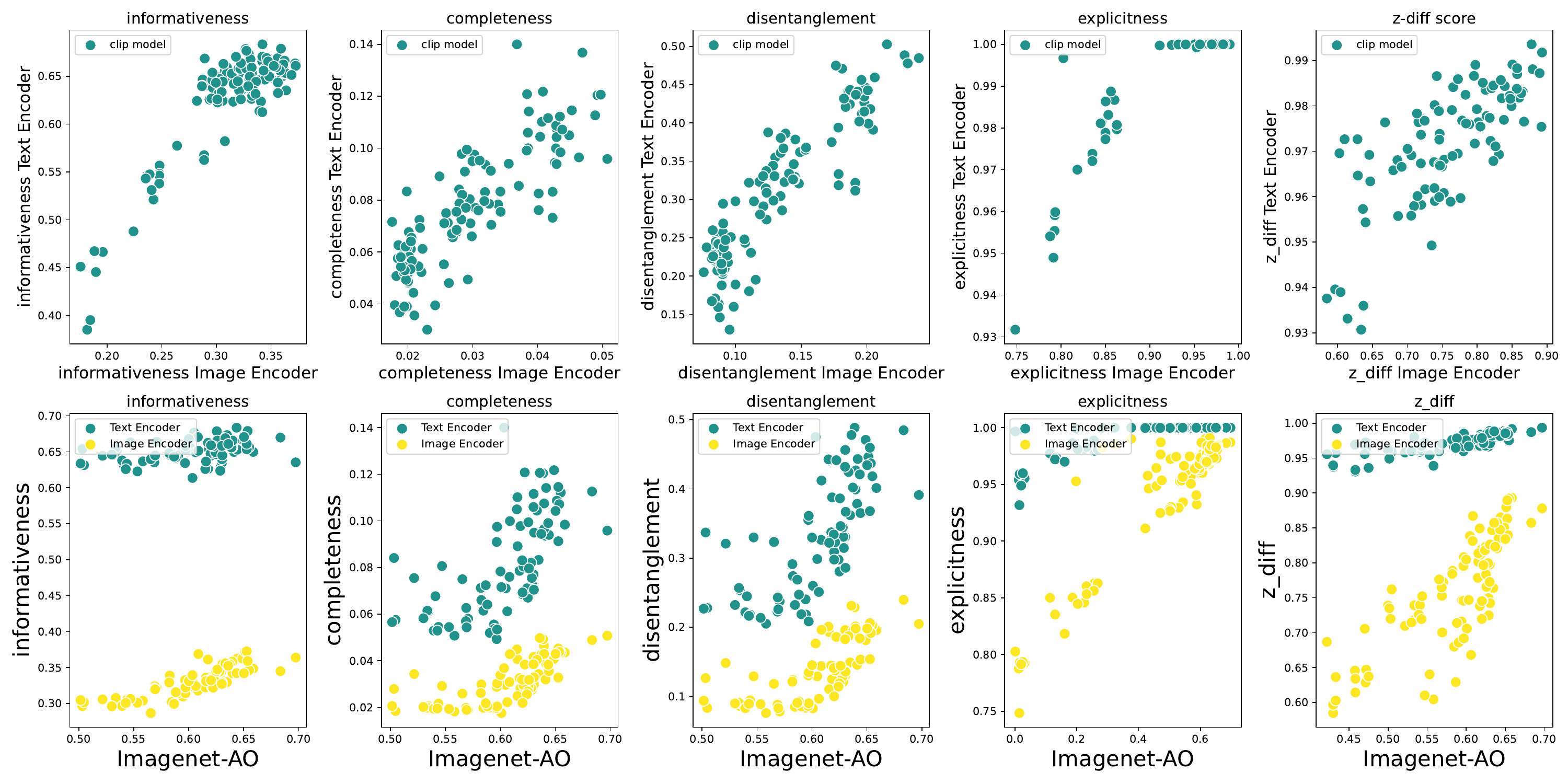} 

  \caption{Top: Representation disentanglments are correlated in text and image embeddings of CLIPs. Bottom: Disentanglment metrics vs. C-OoD Accuracy.}
  
  \label{fig:ood_vs_disentanglement}
\end{figure}

We calculate these metrics for each CLIP model on our ImageNet-AO dataset. Subsequently, in Fig. ~\ref{fig:ood_vs_disentanglement} (bottom), we visualize the relationship between the C-OoD accuracy and the disentanglement metrics. Each point in the plot represents a CLIP model, with the x-axis denoting the C-OoD accuracy and the y-axis representing the disentanglement metric. As observed in bottom row of the plot, there is a discernible pattern where models with higher C-OoD accuracy tend to exhibit more disentangled text and image representations. This empirical observation aligns with our initial hypothesis. Notably, the disentanglement in the text embedding (blue points),  is more pronounced compared to the image embeddings (green points). Additionally, in ~\ref{fig:ood_vs_disentanglement} (top), we show the correlation between the image encoder and the text encoder for different disentanglement metrics. This figure demonstrates that by increasing the disentanglement in the text encoder, the disentanglement in the image encoder also increases, indicating a correlation between them. 

\subsubsection{Intrinsic Dimensionality of the Composition Representations}

\begin{figure}[!t]

  \centering
  \includegraphics[width=\textwidth]{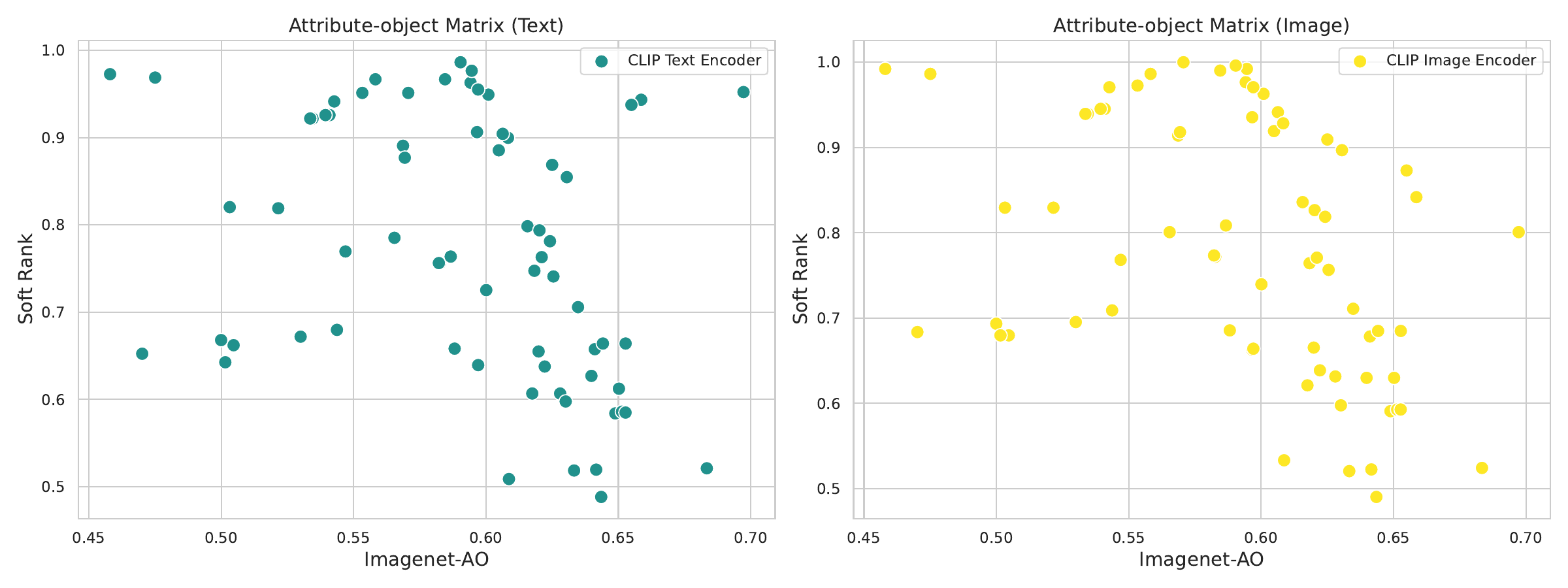} 
    
  \caption{The decrease in the soft rank of attribute-object representations relative to the embedding size correlates with improved C-OoD accuracy. This indicates that decomposing  representations of attributes  and objects  results in a low dimensional representation of CLIPs that exhibits robust C-OoD performance. This highlights the representation disentanglement in CLIPs with strong C-OoD generalization.}
\label{FigSoftRank}
\end{figure}

The previously reported metrics of disentanglement focus on the correspondence between embedding dimensions and latent factors, and hence often require training an auxiliary classifier, in which a given representation is classified into levels of any latent factor. One could alternatively take a training-free approach through measuring relative intrinsic dimensionality of the composition patterns. This could be achieved by measuring the soft rank of the embeddings of attribute-object pairs. The soft rank is defined by the number of singular values of a given matrix that are greater than a pre-specified positive threshold. The soft rank is then normalized and made comparable across CLIPs by being divided to the number of embedding dimensions. This way the soft rank measures the relative intrinsic dimensionality of the embedding space. If the representation is entirely disentangled, huge combinations of attribute-objects would only result in a small intrinsic dimensionality, i.e. sum of the intrinsic dimensionalities of object and attribute spaces. Otherwise, each attribute-object embedding would appear to be {\it novel} with respect to other composition embeddings, resulting in a near full-rank space.

For this experiment, we use ImageNet-AO, which provides around 23,000 unique combinations of attributes and objects. We utilize their image embeddings, obtained from the CLIP image encoder, and caption embeddings, obtained from the CLIP text encoder, to calculate the soft rank with a threshold of 0.1.
Fig. \ref{FigSoftRank} shows that the intrinsic dimensionality is decreasing as the C-OoD accuracy increases, in both text and image domains.

\subsubsection{ Image retrieval with image±text queries}
\begin{figure*}[!t]
  \centering
\includegraphics[width=\textwidth]{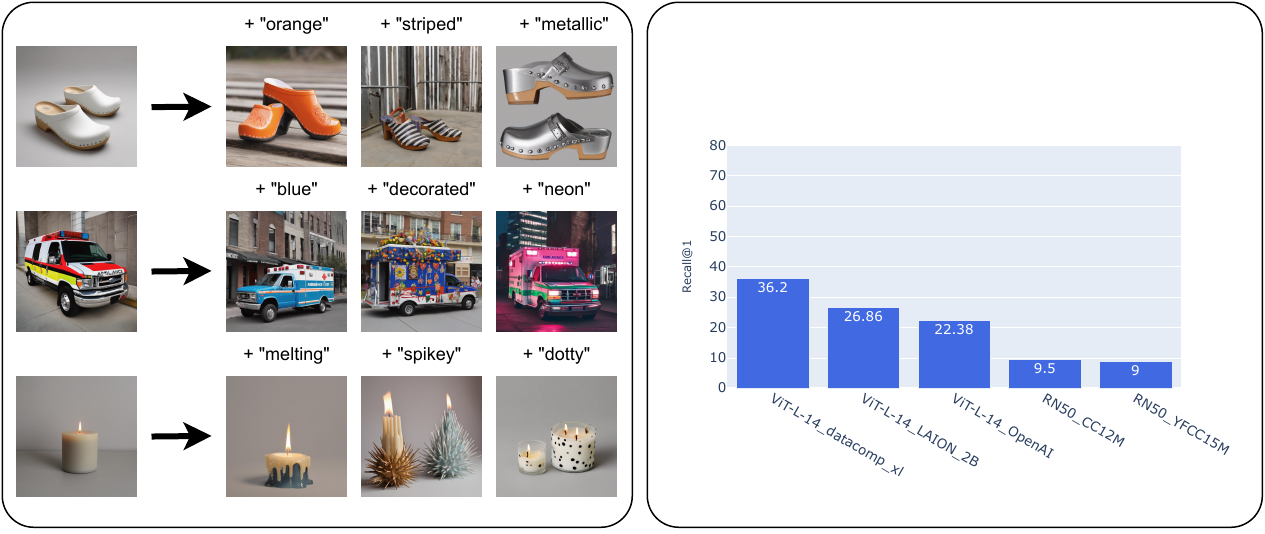} 
  \caption{The performance of various CLIP models in the task of image±text retrieval. A model's superior performance in this task indicates that its representation is more decomposable.}
  \label{fig::Retrieval_disen}
\end{figure*}
Inspired by the work of \cite{jia2021scaling}, we designed an experiment to evaluate the compositional nature of embeddings learned by the CLIP models. Our primary objective is to assess the   representation disentanglement of the CLIP models trained on diverse datasets. To accomplish this goal, we devised a test in which we input an image from our dataset into the image encoder of the model, and obtain its corresponding embedding. Next, we employed the text encoder of the model to compute the embedding of an adjective, ensuring that the adjective differed from those associated with the current image. These two embeddings were then combined through summation and used as a query in a process similar to the image retrieval. We then show the image closest to the generated query embedding. A total of 200 random images were used to conduct this test for each model.

In order to evaluate the accuracy of the models predictions, we consider the image that is most similar to the query as the correct prediction if it possess both the intended object and adjective. A higher level of accuracy in the image retrieval task indicates that the model embeddings are more disentangled. Model evaluations are demonstrated in Fig. \ref{fig::Retrieval_disen}. The Recall@1 performance of  various models aligns with our expectations. Specifically, we anticipated that models excelling in C-OoD tasks would also exhibit more disentangled representations. We previously observed in Fig. \ref{fig:myfigure} that CLIPs associated with LAION and DataComp datasets stand out as having highest C-OoD accuracies. These two CLIPs also performed best in this experiment.

\subsection{ Disentanglement of Fine-Grained Factors}
In the field of Disentanglement Representation Learning, the concept of disentanglement is explored from two distinct perspectives: fine-grained factors at the dimension level and coarse-grained factors at the vector level \cite{wang2023disentangled}. Our initial investigation into CLIP models, utilizing our curated dataset, provided insights into coarse-grained disentanglement (e.g. separating attributes and objects as two factors) and revealed multifaceted evaluation metrics. Moving forward, we aim to delve into the realm of fine-grained disentanglement at the dimension level. However, our current dataset poses inherent limitations in segregating factors at such a granular level. Consequently, to facilitate a comprehensive evaluation of fine-grained disentanglement, it becomes necessary to adopt specialized datasets designed explicitly for disentanglement studies within this domain.



\begin{figure}[!t]
  \centering
  \includegraphics[width=\textwidth]{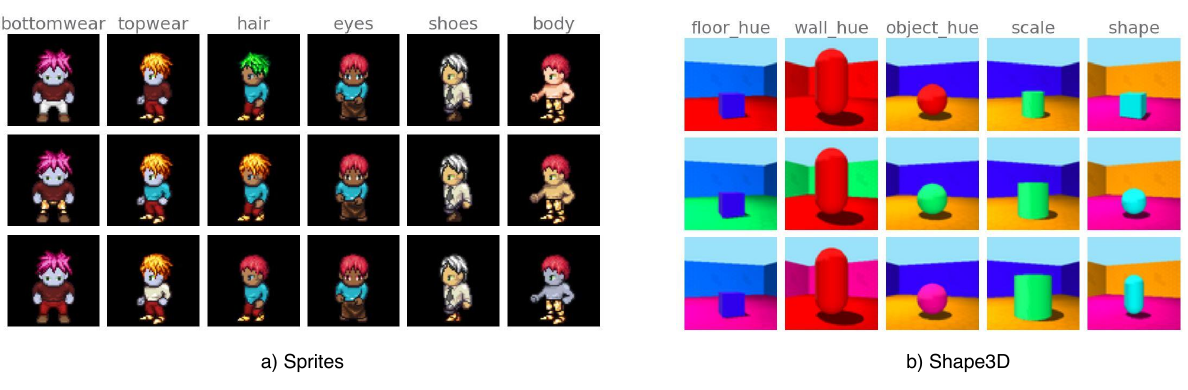} 

  \caption{Disentanglement datasets. a: Sprites dataset, consist of 6 factor and 54,000 images b: Shape3D, consist of 5 factor and 32,000 images }
  
  \label{FineGrainedSamples}
\end{figure}

For our in-depth analysis of the fine-grained disentanglement, we selected two distinguished datasets: Sprites\cite{li2018disentangled}, Shapes3D\cite{3dshapes18} as they are specifically designed for disentanglement studies in image-centric models. Examples from these datasets can be seen in Fig. \ref{FineGrainedSamples}.

Since our focus extends beyond image-centric models to evaluate disentanglement in both the text encoder and image encoder components of CLIP models, we generated captions for each image based on the vector of factors associated with that image. This approach enables us to assess the disentanglement capabilities of CLIP models in both the visual and textual domains.



\begin{figure}[!t]
  \centering
  \includegraphics[width=\textwidth]{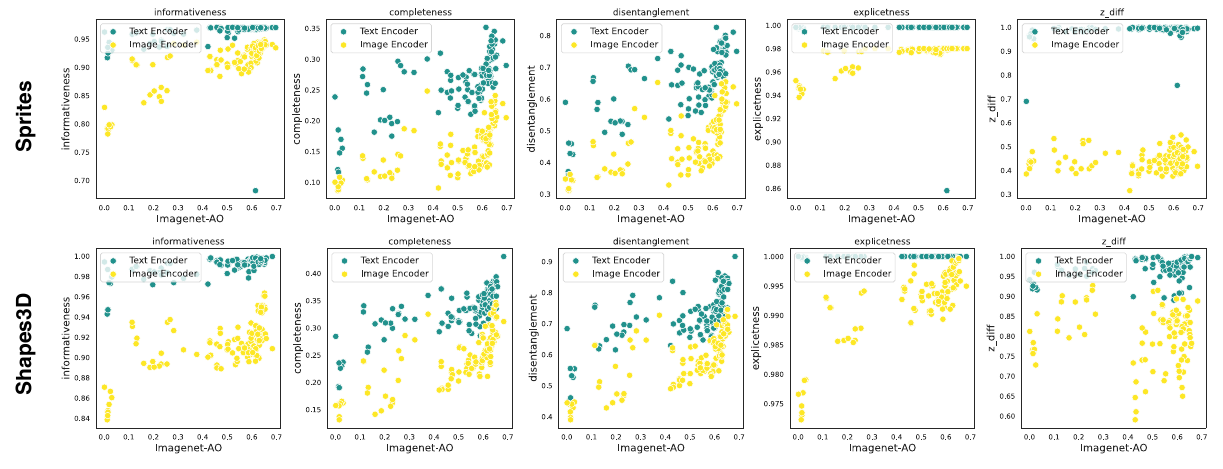} 

  \caption{Disentanglment metrics vs. C-OoD Accuracy on Sprites and Shapes3D dataset.}
  
  \label{fig:sprites-cars3d-metrcis}
\end{figure}

Figure~\ref{fig:sprites-cars3d-metrcis} shows the text encoder exhibits higher disentanglement than the image encoder. As models improve on the C-OoD task, disentanglement tends to increase for both encoders.

\subsubsection{More Analysis on decomposability of the representation space}
Using the Shapes 3D, we conducted two experiments to investigate the representation of factors more accurately. 

In first experiment, we employ the 480,000 images of Shapes 3D dataset, each with specific latent factors such as floor hue, wall hue, object hue, scale, shape and orientation. We train a classifier to calculate the Z-Diff Score and utilize it to determine which dimensions are most critical for each latent factor. In the process of calculating the Z-Diff score, we train a classifier that can determine, for a group of data points that have a fixed specific value for one of the latent factors, what that factor is. By using this classifier, we can identify which dimensions are more important for determining each factor. Subsequently, we extract the top 100 important dimensions for each factor and calculate how many dimensions are common across factors. Our results, presented in Table \ref{tab:common_dimension}, demonstrate that models with higher C-OoD accuracy tend to exhibit fewer common dimensions across factors. This finding suggests that improved C-OoD generalization is associated with more disentangled representations.

In the second experiment, we looked at the impact of disentanglement on zero-shot object color manipulation using two identical images except for the object color. We calculated the embeddings using the CLIP and used the classifier of the first experiment to identify the most important dimensions for detecting object color. By switching the top $k$ dimensions between the two image embeddings, we tested the models' ability to detect captions matching the switched new color. The results are summarized in Table \ref{tab:common_dimension} showing that models with higher C-OoD accuracy require fewer dimension switches to achieve the color change, indicating that  disentangled representations enable more effective zero-shot modifications.



  

\begin{table}[h]
  \centering
  \caption{Number of common dimensions across factors and switching dimensions for color manipulation in the Shapes 3D dataset}
  \vskip 0.01in
  \small 
  \setlength{\tabcolsep}{4pt} 
  \begin{tabular}{lccccc}
  \toprule
  \textbf{Dataset} & \textbf{Architecture} & \textbf{C-OoD Acc.} & \textbf{\# Com. Dims} & \textbf{\# Sw. Dims}\\
  \midrule
  LAION  & ViT-L/14 & 64.61\% & 2 & 40\\
  LAION  & ViT-B/16 & 61.55\% & 5 & 60\\
  LAION  & ViT-B/32 & 61.05\% & 7 & 90\\
  
  OpenAI & ViT-L/14 & 52.28\% & 3 & 5\\
  OpenAI & ViT-B/16 & 49.22\% & 4 & 10\\
  OpenAI & ViT-B/32 & 47.07\% & 6 & 30\\
  
  CC     & RN50     & 26.64\%  & 15 & 200\\
  YFCC   & RN50     & 12.23\%  & 21 & 250\\
  \bottomrule
  \end{tabular}
  \label{tab:common_dimension}
\end{table}

\section{Conclusion}
This study examines how well CLIPs can generalize to new compositions of objects and attributes. We created an authentic benchmark of compositional images that are truly novel with respect to the 
 CLIP training sets, and found that CLIPs ability to decompose the text/images representation space (into the embedding of concepts) is crucial for the compositional generalization. We have assessed the decomposability through the lens of several well-known metrics, as well the composition representation intrinsic dimensionality. These experiments were conducted on a wide range of datasets, from our attribute-object dataset to the ones previously designed specifically to evaluate disentanglement. We also covered a wide variety of problem setups in this direction, ranging from factor classification, and image$\pm$text retrieval, to factor manipulation. All mentioned assessments consistently demonstrate a strong connection between text and image representation disentanglement and C-OoD generalization. 

%
%
\bibliographystyle{unsrt}
\bibliography{main}

\newpage

\section{Appendix}

\subsection{Related Work}
\subsubsection{disentanglement and generalization}

Schott et al. \cite{schott2022visualrepresentationlearningdoes} demonstrated that learning disentangled representations does not inherently lead to strong generalization performance within the same domain. Their findings highlight the difficulty models face in accurately inferring underlying generative factors, even when trained with varying levels of supervision. Montero et al. \cite{montero2020role} explored the role of disentanglement in generalization, focusing on combinatorial generalization. They found that while disentangled representations can enhance interpretability and sample efficiency, they do not necessarily support more complex forms of generalization. This work underscores the intricate relationship between disentanglement and generalization, suggesting that disentangled representations alone are insufficient for achieving advanced generalization capabilities. Further, Montero et al. \cite{montero2022lost} examined the relationship between disentangled representations and combinatorial generalization. They demonstrated that even models with highly disentangled latent spaces often fail to generalize to unseen combinations of generative factors, highlighting the challenges in achieving both disentanglement and robust generalization.

\subsubsection{benchmarks for compositionality in VLMs}

The CREPE benchmark \cite{ma2023crepe} is notable for its introduction of measures of systematicity and productivity to assess how well these models can generalize from known combinations of visual and textual elements to novel compositions. Despite large-scale pretraining, CREPE revealed that these models face significant challenges in compositional reasoning.
Building on CREPE, the Cola benchmark \cite{ray2024cola} was specifically developed to address the limitations of VLMs in compositionality. It focuses on the models' ability to accurately retrieve images based on the correct configuration of objects and their attributes, presenting a challenging testbed for evaluating and improving compositional reasoning in VLMs.
However, critical vulnerabilities and biases in existing benchmarks, including Winoground, VL-CheckList, ARO, CREPE, and Cola, were identified, where blind models often outperformed vision-language models due to hackable biases. Addressing these issues, Hsieh et al. \cite{hsieh2024sugarcrepe} introduced SUGARCREPE, a benchmark designed to evaluate vision-language compositionality by generating fluent and plausible hard negatives using large language models and adversarial refinement. This work aims to provide a more robust and unbiased evaluation of VLMs' compositional abilities.

\subsubsection{ Compositional Generalization}

Wiedemer et al. \cite{wiedemer2024compositional} tackled the challenge of compositional generalization in machine learning, with a focus on vision tasks. Their work explores the theoretical underpinnings of COoD generalization by examining the data-generating processes rather than the data itself. They introduced a framework establishing sufficient conditions for compositional generalization based on the support of the training distribution and model architecture. Their empirical validation demonstrates the practical applicability of their theoretical results, setting the stage for a principled study of compositional generalization across various real-world scenarios. Frady et al. \cite{frady2023learning} delved into the theoretical aspects of compositional OoD generalization for vision tasks. They proposed a model that describes visual scenes using structured symbolic distributed representations, employing Vector Symbolic Architecture (VSA). Their approach trains deep neural networks to output a high-dimensional vector representing the full compositional description of a scene, including attributes such as object identity, position, and color. The model is evaluated on its ability to generalize to unseen digit shapes and scene configurations, revealing its strengths and limitations in handling compositional generalization. This work highlights the challenges and potential solutions for achieving robust compositional generalization in vision tasks, which is crucial for developing more adaptable and resilient neural networks. Additionally, Wiedemer et al. \cite{wiedemer2023provable} investigated the conditions under which compositional generalization can be guaranteed in object-centric representation learning. They framed the problem through the lens of identifiability theory, demonstrating that autoencoders satisfying specific structural assumptions on the decoder and enforcing encoder-decoder consistency can learn object-centric representations that generalize compositionally. This theoretical exploration was validated with experiments on synthetic image data, underscoring the practical relevance of their assumptions. This work contributes to understanding when and how object-centric representations can support compositional generalization, addressing a key gap in the theoretical foundations of compositional generalization in vision tasks.

\subsection{Attributes}
The dataset we designed utilizes the following list of 140 attributes from VAW dataset:

cracked, dilapidated, dry, folded, wet, jagged, moss covered, rough, textured, wrinkled, transparent, clean, dirty, dusty, stained blue plaid, checkered, dotted, floral, lined, red striped, speckled, spotted, striped, arch shaped, arrow shaped, circular, conical, cubed, curved, curly, cylindrical, diamond shaped, domed, heart shaped, octagonal, oval shaped, rectangular, round, rounded, spherical, spiky, spiral, square, triangular, aluminum, asphalt, bamboo, brass, brick, cardboard, cement, ceramic, chocolate, chrome, clay, cloth, cobblestone, concrete, denim, dirt, fabric, fluffy, foamy, furry, glass, granite, gravel, hardwood, iron, jean, khaki, leather, marble, metal, muddy, paper, pebbled, plastic, plush, porcelain, red brick, rocky, rubber, sandy, silk, snowy, stainless steel, steel, stone, straw, stucco, styrofoam, tiled, wicker, wooden, water, colorful, red, pink, purple, green, amber, aqua, beige, black, blond, blue, bluish, bronze, brown, burgundy, fuchsia, golden, gray, green, ivory, maroon, murky, orange, pink, purple, purplish, red, reddish, silver, tan, taupe, teal, terracotta, turquoise, violet, white, yellow
\subsection{Dataset Design}
\label{appendix_2}

The dataset creation process involved two main phases: generation and filtering. Each phase consisted of multiple steps to ensure the final dataset's quality, diversity, and suitability for evaluating compositional understanding in vision-language models.

\subsubsection{Generation Phase}

In the first phase, we aimed to create a diverse set of attribute-object compositions to serve as prompts for image generation. We leveraged two well-established datasets: ImageNet and the Visual Attributes in Wild (VAW) dataset.

\paragraph{\textbf{ImageNet}}
This large-scale dataset contains over 14 million images categorized into 1,000 object classes. Its hierarchical structure and extensive annotation make it a reliable source for identifying distinct object categories.

\paragraph{\textbf{VAW Dataset}}
Designed specifically for attribute-centric image representation, the VAW dataset provides a comprehensive collection of visual attributes. These attributes describe various characteristics, such as colors, materials, and textures, enabling the creation of rich and descriptive prompts.
\\

By combining 1,000 ImageNet object classes with 140 carefully selected attributes from the VAW dataset, we generated 140,000 unique attribute-object compositions. These compositions formed the basis for our image generation prompts, allowing us to explore a wide range of visual concepts.

To generate images from these prompts, we employed the state-of-the-art SDXL-Trubo model \cite{sauer2023adversarial}, a powerful text-to-image generator trained on a vast corpus of image-text pairs. By leveraging the model's ability to translate natural language descriptions into visual representations, we generated approximately 420,000 images corresponding to the attribute-object compositions. To generate the images for our dataset, we employed two main prompt formats:
\begin{itemize}
\item "image of [attribute] [object]"
\item "image of [object] that is [attribute]"
\end{itemize}

\subsubsection{Filtering Phase}

While the generation phase yielded a large initial dataset, further filtering was necessary to ensure reliability, novelty, and adherence to the compositional nature of the prompts. The filtering phase involved several rigorous steps:

\paragraph{\textbf{Composition Validation}}
Text-to-image models can sometimes struggle with accurately depicting compositional concepts, leading to inconsistencies between the prompt and the generated image. To address this issue, we manually inspected each image and removed those where the attribute or object was incorrectly represented or missing entirely.

\paragraph{\textbf{Dataset Novelty}}
To ensure the novelty of our dataset, we searched for compositions present in existing datasets commonly used for training CLIP models: LAION, CommonPool (DataComp), YFCC15m and CC12m. If a composition was found in these datasets, we removed it and its corresponding images from our dataset. During this search, we took a conservative approach: if an attribute and an object existed in the dataset captions, even if not adjacent, we removed the composition from our dataset to avoid any potential overlap.

\paragraph{\textbf{Similarity Filtering}}
Even after removing exact matches, our dataset might still contain images visually similar to those in existing datasets. To address this, we employed Faiss, a library for efficient similarity search. We calculated image embeddings for our dataset and the LAION and CommonPool datasets, then applied the K-Nearest Neighbors (KNN) algorithm to identify highly similar images. Compositions with highly similar counterparts in these datasets were removed from our dataset, further ensuring its novelty and distinctiveness.

After applying these rigorous filtering steps, the remaining images and compositions constitute an out-of-distribution dataset for CLIP models. This final dataset represents a diverse and reliable collection of attribute-object combinations, carefully curated to evaluate the compositional understanding capabilities of vision-language models in a novel and challenging setting.

\begin{table}[ht]
\centering
\caption{Comparison Before and After Filtering}
\label{tab:comparison}
\begin{tabular}{lcc}
\toprule
 & \textbf{Initial Dataset} & \textbf{Final Dataset} \\
\midrule
Attribute & 140 & 87 \\
Object & 1000 & 663 \\
Attribute-Object & 140,000 & 20,364 \\
\bottomrule
\end{tabular}
\end{table}

\begin{figure}[!t]
  \centering
  \includegraphics[width=\textwidth]{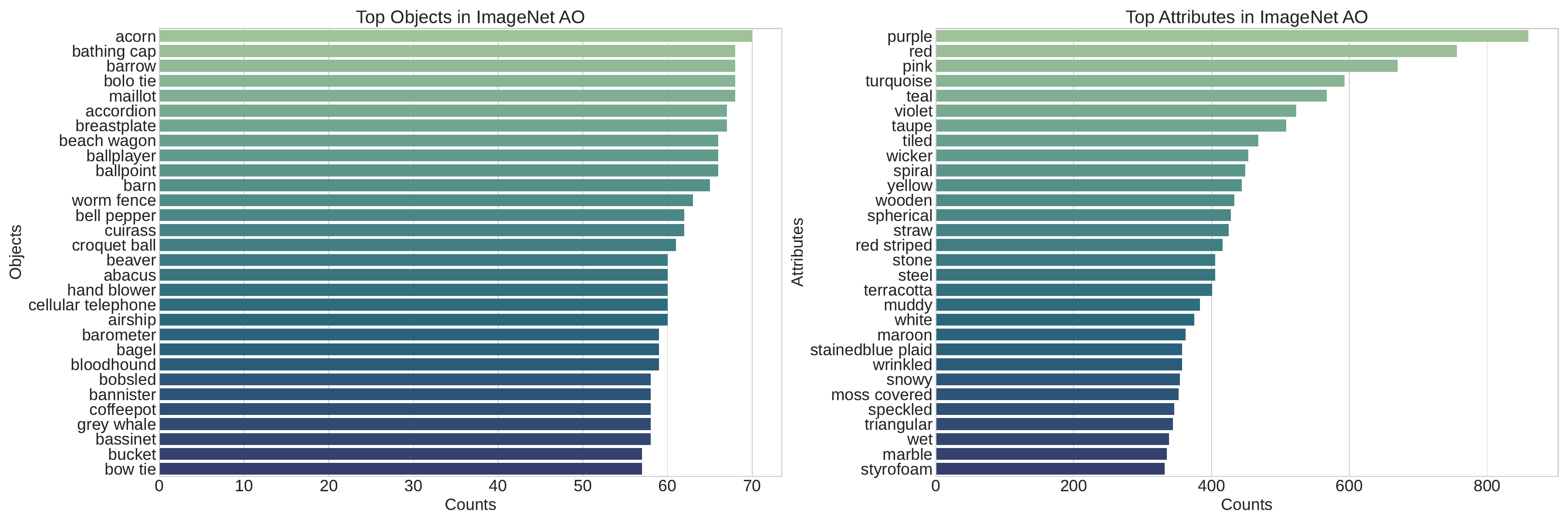} 

  \caption{Attributes and objects with highest frequencies in the ImageNet-AO dataset.
}
  
  \label{}
\end{figure}

\subsection{Zero shot Evaluation on ImageNet-AO}
\label{appendix_1}
In this section, we report the detailed results of various models on the ImageNet-AO dataset.

{\footnotesize 
\begin{table}[t]
\centering
\caption{Performance comparison on ImageNet-AO (Part 1).}
\label{tab:results_part1}
\adjustbox{max width=\textwidth}{ 
\begin{tabular}{llr}
\toprule
\textbf{Name} & \textbf{Pretrained} & \textbf{Imagenet-AO} \\
\midrule
EVA02-E-14-plus & laion2b\_s9b\_b144k & 70.88 \\
ViT-bigG-14 & laion2b\_s39b\_b160k & 69.88 \\
EVA02-E-14 & laion2b\_s4b\_b115k & 68.29 \\
convnext\_xxlarge & laion2b\_s34b\_b82k\_augreg\_soup & 67.58 \\
convnext\_xxlarge & laion2b\_s34b\_b82k\_augreg\_rewind & 67.39 \\
convnext\_xxlarge & laion2b\_s34b\_b82k\_augreg & 67.25 \\
convnext\_large\_d\_320 & laion2b\_s29b\_b131k\_ft\_soup & 66.94 \\
EVA01-g-14-plus & merged2b\_s11b\_b114k & 66.81 \\
ViT-H-14-CLIPA-336 & laion2b & 66.53 \\
ViT-H-14 & laion2b\_s32b\_b79k & 66.50 \\
ViT-g-14 & laion2b\_s12b\_b42k & 66.45 \\
convnext\_large\_d\_320 & laion2b\_s29b\_b131k\_ft & 66.33 \\
convnext\_large\_d & laion2b\_s26b\_b102k\_augreg & 66.25 \\
ViT-H-14-378-quickgelu & dfn5b & 66.15 \\
ViT-bigG-14-CLIPA & datacomp1b & 65.53 \\
ViT-bigG-14-CLIPA-336 & datacomp1b & 65.46 \\
ViT-H-14-CLIPA & datacomp1b & 65.43 \\
ViT-H-14-quickgelu & dfn5b & 65.38 \\
ViT-H-14-CLIPA-336 & datacomp1b & 65.18 \\
ViT-SO400M-14-SigLIP-384 & webli & 65.18 \\
ViT-B-16-SigLIP-384 & webli & 64.58 \\
ViT-B-16-SigLIP-256 & webli & 64.55 \\
convnext\_base\_w\_320 & laion\_aesthetic\_s13b\_b82k\_augreg & 64.33 \\
ViT-H-14 & laion2b\_s32b\_b79k & 64.15 \\
ViT-B-16-SigLIP-512 & webli & 64.11 \\
coca\_ViT-L-14 & laion2b\_s13b\_b90k & 63.80 \\
ViT-L-14-CLIPA-336 & datacomp1b & 63.61 \\
ViT-L-14 & commonpool\_xl\_clip\_s13b\_b90k & 63.55 \\
convnext\_base\_w & laion2b\_s13b\_b82k\_augreg & 63.50 \\
convnext\_base\_w\_320 & laion\_aesthetic\_s13b\_b82k & 63.42 \\
ViT-H-14-quickgelu & metaclip\_fullcc & 63.22 \\
ViT-L-14-quickgelu & metaclip\_fullcc & 63.10 \\
convnext\_base\_w & laion2b\_s13b\_b82k & 63.02 \\
ViT-B-16-SigLIP-i18n-256 & webli & 63.02 \\
ViT-L-14-quickgelu & dfn2b & 62.99 \\
ViT-L-14 & datacomp\_xl\_s13b\_b90k & 62.83 \\
ViT-B-16-SigLIP & webli & 62.76 \\
coca\_ViT-L-14 & mscoco\_finetuned\_laion2b\_s13b\_b90k & 62.75 \\
convnext\_base\_w & laion\_aesthetic\_s13b\_b82k & 62.57 \\
ViT-L-16-SigLIP-256 & webli & 62.55 \\
\bottomrule
\end{tabular}
}
\end{table}
}

\begin{table}[htbp]
\centering
\caption{Performance comparison on ImageNet-AO (Part 2).}
\label{tab:results_part2}
\adjustbox{max width=\textwidth}{ 
\begin{tabular}{llr}
\toprule
\textbf{Name} & \textbf{Pretrained} & \textbf{Imagenet-AO} \\
\midrule
ViT-SO400M-14-SigLIP & webli & 62.50 \\
EVA01-g-14 & laion400m\_s11b\_b41k & 61.87 \\
EVA02-L-14-336 & merged2b\_s6b\_b61k & 61.83 \\
ViT-B-16 & datacomp\_xl\_s13b\_b90k & 61.63 \\
EVA02-L-14 & merged2b\_s4b\_b131k & 61.48 \\
ViT-L-16-SigLIP-384 & webli & 61.34 \\
ViT-B-16 & laion2b\_s34b\_b88k & 61.09 \\
ViT-L-14 & commonpool\_xl\_laion\_s13b\_b90k & 61.01 \\
ViT-L-14 & laion400m\_e32 & 60.70 \\
ViT-B-16-quickgelu & metaclip\_fullcc & 60.70 \\
ViT-L-14 & laion400m\_e31 & 60.67 \\
ViT-B-16 & dfn2b & 60.66 \\
ViT-B-32 & laion2b\_s34b\_b79k & 60.46 \\
ViT-B-32 & laion2b\_e16 & 60.42 \\
ViT-B-32-256 & datacomp\_s34b\_b86k & 60.04 \\
roberta-ViT-B-32 & laion2b\_s12b\_b32k & 59.82 \\
xlm-roberta-base-ViT-B-32 & laion5b\_s13b\_b90k & 59.67 \\
ViT-B-32 & datacomp\_xl\_s13b\_b90k & 59.45 \\
ViT-B-16-plus-240 & laion400m\_e31 & 59.27 \\
ViT-B-16-plus-240 & laion400m\_e32 & 59.22 \\
EVA02-B-16 & merged2b\_s8b\_b131k & 58.06 \\
coca\_ViT-B-32 & laion2b\_s13b\_b90k & 57.93 \\
ViT-B-16 & laion400m\_e31 & 57.89 \\
ViT-B-16 & laion400m\_e32 & 57.87 \\
ViT-L-14-quickgelu & metaclip\_400m & 57.54 \\
ViT-B-32-quickgelu & metaclip\_fullcc & 56.82 \\
ViT-B-16-quickgelu & metaclip\_400m & 56.33 \\
ViT-L-14 & commonpool\_xl\_s13b\_b90k & 54.69 \\
ViT-B-16 & datacomp\_l\_s1b\_b8k & 54.37 \\
convnext\_base & laion400m\_s13b\_b51k & 54.27 \\
ViT-B-32-quickgelu & laion400m\_e31 & 54.09 \\
ViT-B-32-quickgelu & laion400m\_e32 & 53.94 \\
ViT-B-32 & laion400m\_e31 & 53.45 \\
ViT-B-32 & laion400m\_e32 & 53.36 \\
ViT-B-16 & commonpool\_l\_clip\_s1b\_b8k & 53.00 \\
ViT-L-14-336 & openai & 52.15 \\
ViT-B-16 & commonpool\_l\_laion\_s1b\_b8k & 50.47 \\
ViT-L-14 & openai & 50.32 \\
ViT-B-16 & commonpool\_l\_text\_s1b\_b8k & 50.15 \\
ViT-B-16 & commonpool\_l\_image\_s1b\_b8k & 50.00 \\
ViT-B-16 & openai & 47.51 \\
RN50x64 & openai & 47.19 \\
RN50x16 & openai & 47.13 \\
ViT-B-16 & commonpool\_l\_basic\_s1b\_b8k & 47.02 \\
ViT-B-32-quickgelu & openai & 45.80 \\
ViT-B-32 & openai & 45.80 \\
RN50x4 & openai & 45.75 \\
\bottomrule
\end{tabular}
}
\end{table}


  

\subsection{Disentanglement measures}
\label{appendix_3}

\subsubsection{Disentanglement:}

The disentanglement metric quantifies the degree to which the learned representation factorizes or disentangles the underlying generative factors of variation. Ideally, each dimension (or variable) of the learned representation should capture at most one generative factor. The disentanglement score $D_i$ for a code variable $c_i$ is defined as:

\begin{equation}
D_i = 1 - H_K(P_{i\cdot})
\end{equation}
where $H_K(P_{i\cdot}) = -\sum_{k=0}^{K-1} P_{ik} \log_K P_{ik}$ is the entropy, and $P_{ij} = R_{ij} / \sum_{k=0}^{K-1} R_{ik}$ represents the relative "importance" of $c_i$ for predicting the generative factor $z_j$. If $c_i$ is important for predicting a single generative factor, its disentanglement score $D_i$ will be 1. If $c_i$ is equally important for predicting all generative factors, its score will be 0.



\subsubsection{Completeness:}

The completeness metric quantifies the degree to which each underlying generative factor is captured by a single code variable. The completeness score $C_j$ for capturing the generative factor $z_j$ is defined as:

\begin{equation}
C_j = 1 - H_D(\tilde{P}_{.j})
\end{equation}
where $H_D(\tilde{P}_{.j}) = -\sum_{d=0}^{D-1} \tilde{P}_{dj} \log_D \tilde{P}_{dj}$ denotes the entropy of the $\tilde{P}_{.j}$ distribution. If a single code variable contributes to the prediction of $z_j$, the completeness score $C_j$ will be 1 (complete). If all code variables equally contribute to the prediction of $z_j$, the score will be 0 (maximally overcomplete).

The completeness score $C_j$ quantifies how well the generative factor $z_j$ is captured by a single code variable in the learned representation. A higher score indicates that the factor is more completely represented by a single variable, without being overcomplete (i.e., requiring multiple variables to represent the factor).

\subsubsection{Informativeness:}

The informativeness metric quantifies the amount of information that a representation captures about the underlying factors of variation. The informativeness of the code $c$ about the generative factor $z_j$ is quantified by the prediction error $E(z_j, \hat{z}_j)$ (averaged over the dataset), where $E$ is an appropriate error function and $\hat{z}_j = f_j(c)$.

It is important to note that the prediction error $E(z_j, \hat{z}_j)$, and thus this informativeness metric, is dependent on the capacity of $f$, with linear regressors only capable of extracting information about $z$ in $c$ that is explicitly represented. Hence, this informativeness metric is also dependent on a model's ability to explicitly represent information about $z$ in $c$, which in turn is dependent on the model's ability to disentangle the underlying factors of variation ($z$). Thus, the informativeness metric has some overlap with the disentanglement metric, with the size of the overlap determined by the capacity of $f$ (no overlap with infinite capacity).

The informativeness metric quantifies how much information about the generative factors is captured in the learned representation, with lower prediction errors indicating higher informativeness. Representations that are highly informative about the underlying factors are desirable for tasks that require knowledge of the important attributes of the data.

\subsubsection{Z-diff:}

The Z-diff metric, also known as the $\beta$-VAE metric, is a disentanglement metric that evaluates the learned representation by training a linear classifier to predict which generative factor was held constant between pairs of instances. The metric works as follows:

1. Pairs of instances are selected to create batches. In each batch, a factor $v_i$ is chosen randomly.

2. A fixed number of pairs are formed with samples $v_1$ and $v_2$ that have the same value for the chosen factor ($v_{1_i} = v_{2_i}$).


3. Each pair is represented by the absolute difference of the codes associated with the samples ($p = z_1 - z_2$).
4. The intuition is that code dimensions associated with the fixed factor should have the same value, resulting in a smaller difference than the other code dimensions.
5. The mean of all pair differences in the subset creates a point in a final training set.
6. This process is repeated several times to constitute a sizable training set.
7. Finally, a linear classifier is trained on the data set to predict which factor was fixed.

The accuracy of the linear classifier on this task is the Z-diff score. For a completely random classifier, we expect an accuracy of $1/M$, where $M$ is the number of generative factors. This can be used to scale the output closer to the $[0, 1]$ range.

The Z-diff metric quantifies how well the learned representation disentangles the generative factors by evaluating the ability of a linear classifier to predict which factor was held constant between pairs of instances based on the difference in their representations. Higher scores indicate better disentanglement of the factors in the learned representation.

\subsubsection{Explicitness:}

The explicitness score is a metric proposed to evaluate how explicitly the generative factors are represented in the learned representation. It assumes that the generative factors have discrete values and uses a classifier trained on the entire representation to predict the factor classes. The metric is computed as follows: 1) A classifier, such as logistic regression, is trained on the representation to predict the factor classes for each generative factor.
2) The classification performance is reported using the area under the receiver operating characteristic curve (AUC-ROC).
3) The final explicitness score is the average AUC-ROC over all classes for all generative factors.

The AUC-ROC has a minimal value of 0.5, which corresponds to a random classifier. To obtain a score in the range of [0, 1], the AUC-ROC values need to be normalized as:

\begin{equation}
\text{Explicitness Score} = \frac{1}{M} \sum_{j=1}^{M} \frac{\text{AUC-ROC}_j - 0.5}{0.5}
\end{equation}
where $M$ is the number of generative factors, and $\text{AUC-ROC}_j$ is the AUC-ROC for the classifier predicting the $j$-th factor.

In the implementation, class weights in the logistic regression loss are balanced to account for class imbalance.

The explicitness score quantifies how explicitly the generative factors are represented in the learned latent code. Higher scores indicate that the factor classes can be more easily predicted from the latent code, suggesting that the factors are more explicitly represented in the learned representation.

\subsection{Correlation Between Disentanglement Metrics and ImageNet Accuracy}

Table \ref{tab:per} presents the correlation coefficients between various disentanglement metrics and ImageNet accuracy for different CLIP models. This table highlights the intricate relationship between the level of disentanglement achieved by a model and its performance on diverse datasets.

\begin{table}[t]
    \centering
    \begin{tabular}{lccccc}
        \toprule
        Dataset & Informativeness & Disentanglement & Completeness & Explicitness & Z-Diff Score \\
        \midrule
        ImageNet-AO & 0.8204 & 0.7645 & 0.7556 & 0.7875 & 0.8412 \\
        Shape3ed & 0.6159 & 0.7792 & 0.8294 & 0.7337 & 0.2150 \\
        dSprites & 0.7644 & 0.8058 & 0.8588 & 0.7515 & 0.2442 \\
        \bottomrule
    \end{tabular}
    \caption{Pearson Correlation Between Disentanglement Metrics and ImageNet Accuracy for Various CLIP Models}
    \label{tab:per}
\end{table}

\subsection{Zero shot Evaluation on variants of the ImageNet dataset}
\label{appendix_4}
\begin{figure}[!t]
  \centering
  \includegraphics[width=\textwidth]{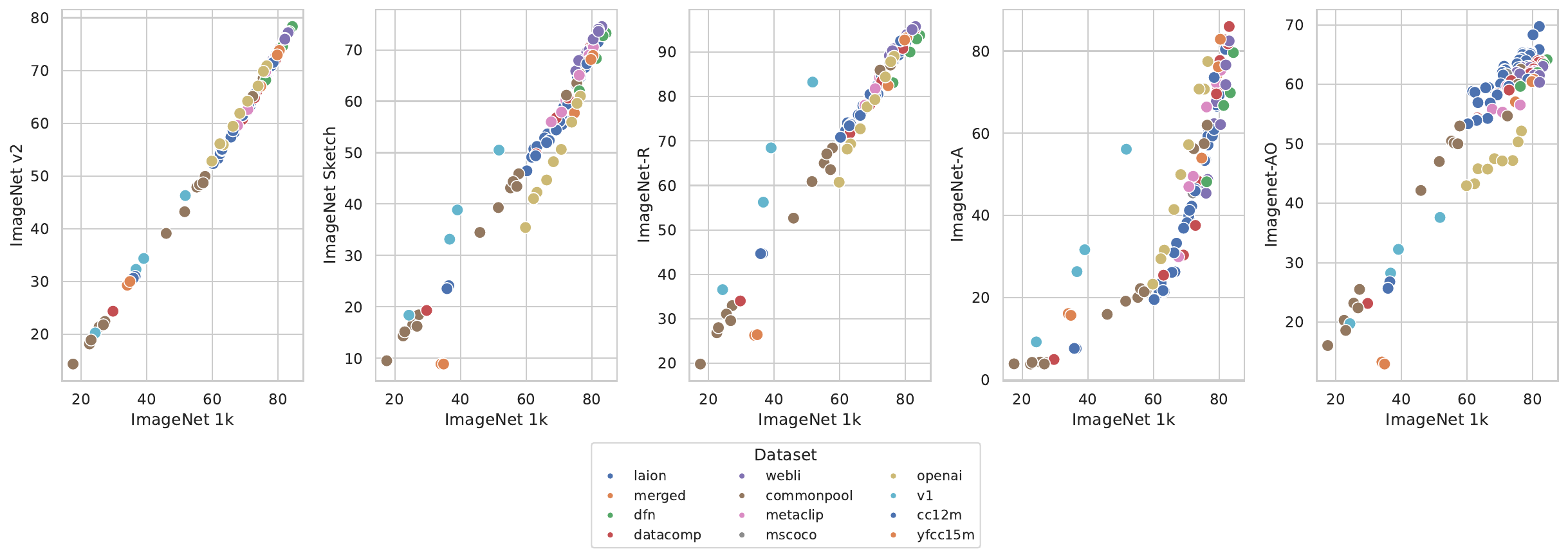} 

  \caption{Performance of various CLIP models on versions of ImageNet with different domain shits vs. in-distribution ImageNet. 
}
  
  \label{fig:imagenet_vs_other}
\end{figure}

\begin{figure}[!t]
  \centering
  \includegraphics[width=0.8\textwidth]{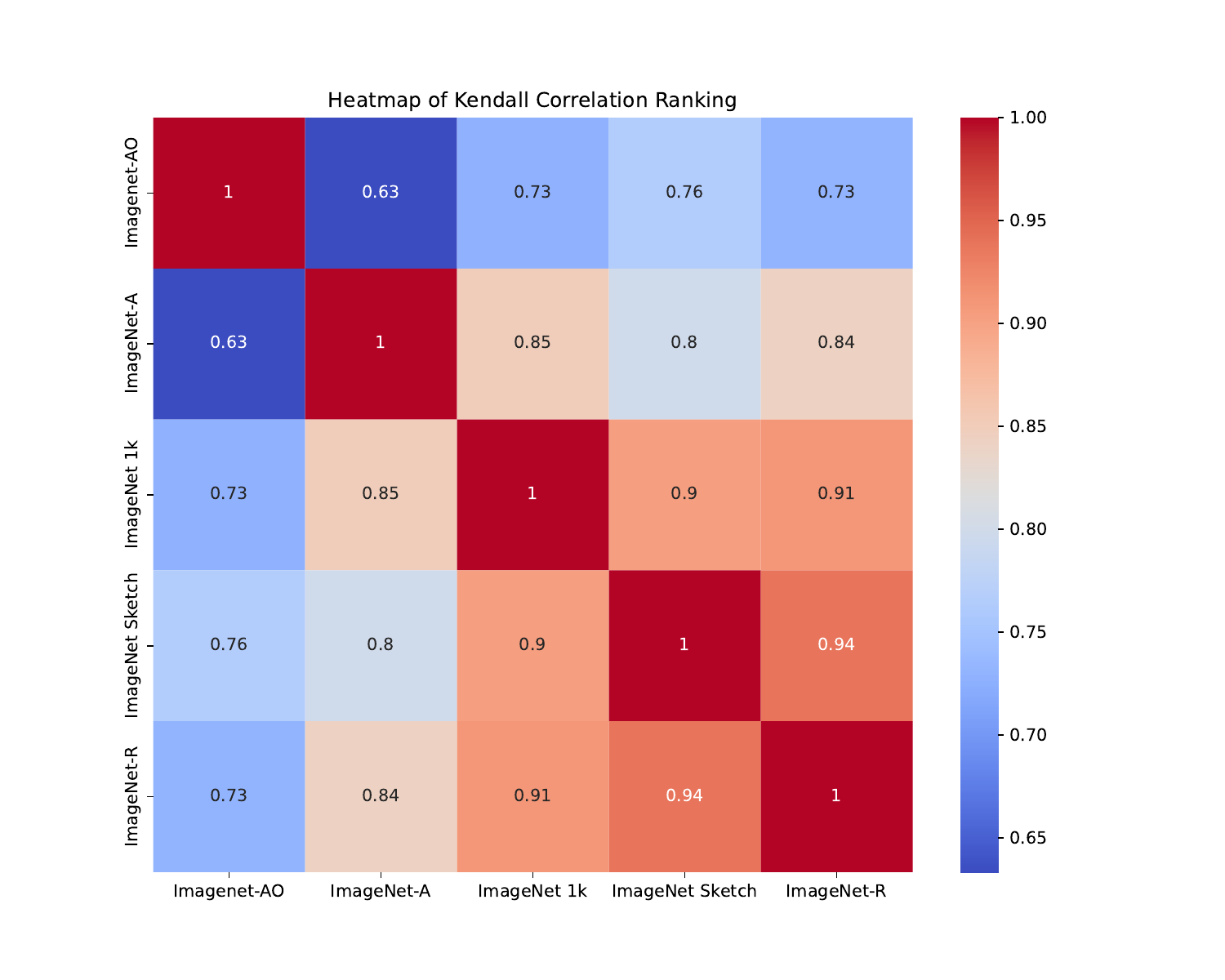} 

  \caption{Kendall rank correlation on different dataset.
}
  
  \label{fig:kendal}
\end{figure}

We evaluate various CLIP models on different versions of the ImageNet dataset, including ImageNetV2, ImageNet-Sketch, ImageNet-R, and ImageNet-A. Our goal is to analyze the performance trends of models on these variant datasets and examine whether they correlate with results on our generated Imagenet-AO dataset. These results are shown in Table \ref{tab:domain_table}.
The scatter plots in Figure \ref{fig:imagenet_vs_other} compares the performance of CLIP models on the ImageNet-1k dataset (x-axis) against their performance on various ImageNet variants (y-axis), providing a visual representation of the correlation between model performance on the original ImageNet and its variants.
 Moreover, for each pair of
the datasets (i.e., domain shifts), the Kendall rank correlation between results of different models
on the corresponding datasets are presented in Figure \ref{fig:kendal}. 
\begin{table}[t]
  \centering
  \caption{Performance on a set of CLIP models on datasets showing various domain shift on ImageNet}
  \label{tab:domain_table}
  \adjustbox{max width=\textwidth}{ 
    \begin{tabular}{lccccccc}
      \toprule
          \textbf{Model} & \textbf{Dataset} & \textbf{ImageNet} & \textbf{ImageNet-v2} & \textbf{Imagenet-sketch} & \textbf{ImageNet-R} & \textbf{ImageNet-A} & \textbf{Imagenet-AO} \\
    \midrule
EVA02-E-14-plus & laion & 82.01 & 75.64 & 71.62 & 94.56 & 82.23 & 70.88 \\
ViT-bigG-14 & laion & 80.09 & 73.59 & 68.94 & 92.13 & 69.33 & 69.88 \\
EVA02-E-14 & laion & 81.96 & 75.66 & 71.51 & 94.07 & 80.44 & 68.29 \\
EVA01-g-14-plus & merged & 79.33 & 72.14 & 68.14 & 92.46 & 74.16 & 66.81 \\
ViT-H-14-CLIPA-336 & laion & 79.10 & 72.41 & 69.94 & 92.69 & 72.13 & 65.18 \\
convnext-xxlarge & laion & 79.47 & 72.60 & 68.40 & 91.60 & 67.19 & 67.25 \\
convnext-large-d-320 & laion & 76.85 & 69.44 & 65.04 & 88.62 & 60.44 & 66.33 \\
convnext-xxlarge & laion & 79.31 & 72.28 & 68.25 & 91.39 & 66.57 & 67.25 \\
xlm-roberta-large-ViT-H-14 & frozen & 76.95 & 69.44 & 65.81 & 89.40 & 59.35 & 65.02 \\
convnext-xxlarge & laion & 79.07 & 72.23 & 68.06 & 91.31 & 66.92 & 67.25 \\
convnext-large-d-320 & laion & 76.60 & 69.29 & 64.72 & 88.23 & 59.33 & 66.33 \\
ViT-g-14 & laion & 76.63 & 69.56 & 65.16 & 88.69 & 57.16 & 65.0 \\
ViT-H-14-378-quickgelu & dfn & 84.37 & 78.33 & 73.24 & 93.76 & 79.64 & 66.15 \\
convnext-large-d & laion & 75.91 & 68.26 & 64.30 & 87.67 & 53.52 & 66.25 \\
ViT-H-14 & laion & 77.96 & 70.90 & 66.57 & 89.34 & 59.35 & 66.5 \\
ViT-H-14-CLIPA & datacomp & 81.52 & 74.98 & 72.72 & 94.26 & 77.01 & 65.43 \\
ViT-bigG-14-CLIPA & datacomp & 82.70 & 76.99 & 74.31 & 95.12 & 81.79 & 65.53 \\
ViT-H-14-CLIPA-336 & datacomp & 81.80 & 75.62 & 72.82 & 94.38 & 82.75 & 65.18 \\
ViT-bigG-14-CLIPA-336 & datacomp & 83.09 & 77.26 & 74.54 & 95.35 & 85.99 & 65.46 \\
ViT-L-14 & laion & 75.25 & 67.80 & 63.28 & 87.42 & 53.85 & 52.29 \\
ViT-H-14-quickgelu & dfn & 83.44 & 77.36 & 72.74 & 92.96 & 69.87 & 63.84 \\
ViT-L-14-CLIPA & datacomp & 79.57 & 73.05 & 70.61 & 92.88 & 71.25 & 64.7 \\
convnext-base-w-320 & laion & 71.28 & 63.62 & 56.46 & 81.36 & 41.57 & 64.03 \\
ViT-SO400M-14-SigLIP-384 & webli & 83.08 & 77.17 & 74.54 & 95.75 & 82.47 & 65.18 \\
ViT-B-16-SigLIP-384 & webli & 78.49 & 72.11 & 69.55 & 92.14 & 62.33 & 64.69 \\
ViT-g-14 & laion & 78.47 & 71.58 & 67.54 & 90.20 & 60.92 & 65.0 \\
ViT-B-16-SigLIP-256 & webli & 76.53 & 69.20 & 68.10 & 90.76 & 48.77 & 64.2 \\
ViT-B-16-SigLIP-512 & webli & 79.14 & 72.83 & 69.90 & 92.64 & 67.69 & 64.45 \\
coca-ViT-L-14 & laion & 75.61 & 67.98 & 64.53 & 88.12 & 53.36 & 64.12 \\
ViT-L-14-CLIPA-336 & datacomp & 80.26 & 73.46 & 70.87 & 93.29 & 77.71 & 64.17 \\
    \bottomrule
    \end{tabular}
  }
\end{table}

\subsection{Text-to-Image Retrieval Evaluation of CLIPs}
\label{sec::t2i}
In this section, we delve into the text-to-image retrieval task and present a thorough evaluation of various CLIP models on ImageNet-AO. The objective of this evaluation is to examine how effectively each CLIP variant can retrieve relevant images based on textual queries, showcasing their ability to bridge the modal gap between language and vision.These results are shown in Table \ref{tab:text2image}.

\begin{table}[ht]
\centering
\caption{zero shot Text-to-Image Performance}
\label{tab:text2image}
\begin{tabular}{cllr}
\toprule
\textbf{ID} & \textbf{Name} & \textbf{Pretrained} & \textbf{R@1} \\
\midrule
99 & EVA02-E-14-plus & laion2b-s9b-b144k & 63.31 \\
73 & ViT-bigG-14 & laion2b-s39b-b160k & 61.70 \\
98 & EVA02-E-14 & laion2b-s4b-b115k & 58.98 \\
88 & convnext-xxlarge & laion2b-s34b-b82k-augreg-soup & 58.21 \\
94 & EVA01-g-14-plus & merged2b-s11b-b114k & 58.05 \\
86 & convnext-xxlarge & laion2b-s34b-b82k-augreg & 58.04 \\
87 & convnext-xxlarge & laion2b-s34b-b82k-augreg-rewind & 57.84 \\
85 & convnext-large-d-320 & laion2b-s29b-b131k-ft-soup & 57.80 \\
84 & convnext-large-d-320 & laion2b-s29b-b131k-ft & 57.44 \\
67 & ViT-H-14 & laion2b-s32b-b79k & 57.41 \\
83 & convnext-large-d & laion2b-s26b-b102k-augreg & 57.12 \\
112 & ViT-H-14-CLIPA-336 & laion2b & 57.04 \\
71 & ViT-g-14 & laion2b-s12b-b42k & 57.01 \\
70 & ViT-H-14-378-quickgelu & dfn5b & 56.66 \\
113 & ViT-H-14-CLIPA-336 & datacomp1b & 56.18 \\
93 & EVA01-g-14 & laion400m-s11b-b41k & 56.16 \\
111 & ViT-H-14-CLIPA & datacomp1b & 55.87 \\
76 & xlm-roberta-large-ViT-H-14 & frozen-laion5b-s13b-b90k & 55.73 \\
58 & ViT-L-14 & laion2b-s32b-b82k & 55.66 \\
115 & ViT-bigG-14-CLIPA-336 & datacomp1b & 55.65 \\
59 & ViT-L-14 & datacomp-xl-s13b-b90k & 55.46 \\
109 & ViT-L-14-CLIPA & datacomp1b & 55.37 \\
72 & ViT-g-14 & laion2b-s34b-b88k & 55.36 \\
110 & ViT-L-14-CLIPA-336 & datacomp1b & 55.34 \\
69 & ViT-H-14-quickgelu & dfn5b & 55.25 \\
103 & ViT-B-16-SigLIP-384 & webli & 55.16 \\
114 & ViT-bigG-14-CLIPA & datacomp1b & 55.03 \\
108 & ViT-SO400M-14-SigLIP-384 & webli & 54.98 \\
60 & ViT-L-14 & commonpool-xl-clip-s13b-b90k & 54.91 \\
101 & ViT-B-16-SigLIP-256 & webli & 54.89 \\
\bottomrule
\end{tabular}
\end{table}

\subsection{Analyzing Compositionality in CLIP Text and Image Representations}

\begin{figure}[!t]
  \centering
  \includegraphics[width=\textwidth]{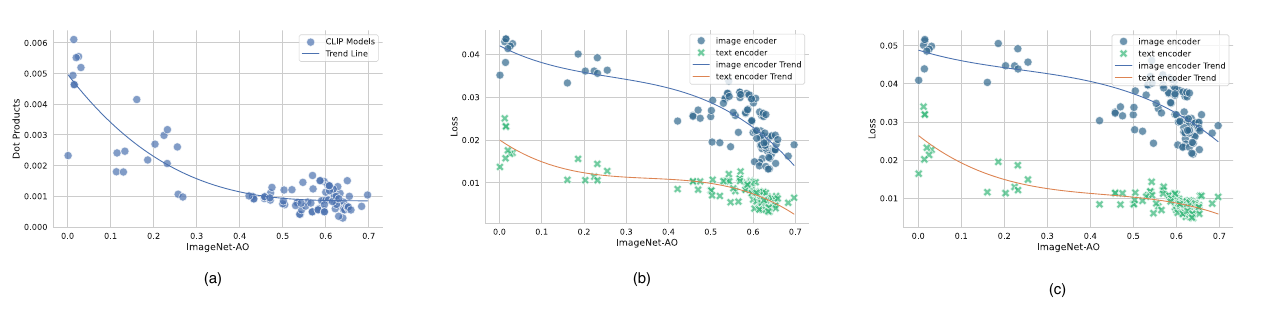} 

  \caption{(a) Orthogonality between adjective and noun representations in the CLIP text encoder, measured by the dot product between their value vectors. Models with higher ImageNet-AO accuracy exhibit greater orthogonality.(b) Reconstruction loss for predicting component (adjective and noun) embeddings from their combined embedding using a single-layer network. Lower loss values indicate better preservation of component information in the combined embedding. The loss is shown for both the text encoder (green points) and image encoder (blue points) of CLIP models, with models having higher ImageNet-AO accuracy tending to have lower reconstruction loss. The text encoder exhibits lower reconstruction loss compared to the image encoder. (c) Same as (b) but using a two-layer network where the second layer's weights are shared and transposed from the first layer. The trend is similar, with higher ImageNet-AO accuracy models having lower reconstruction loss.
}
  
  \label{fig:orth}
\end{figure}

In this section, we designed two experiments to investigate whether the representations of the composition constituents are near-orthogonal, which in turns help the compositional behavior of the CLIP models.

In the first experiment, we focused on the text representations. We fed attribute-object combinations as input to the CLIP text encoder and extracted the value vectors for the object and attribute tokens from the final layer. The value vectors act as the basis of the embeddings in the attention mechanism. In case that such basis members are orthogonal, one could easily decompose the final CLS embedding of the composition into the embeddings of the object and attributes, facilitating the compositional generalization. In order to assess this orthogonality, 
we calculated the dot product between the two value vectors. As depicted in Figure \ref{fig:orth}.a, models with higher accuracy on the ImageNet-AO dataset exhibit higher orthogonality between the value vectors of the  attribute and object.

Having observed the near-orthogonality of the value vectors, for the second experiment,  we try to make the mentioned decomposition of the CLS token of the composition into its constituents. An important difference of this experiment to the last one is that the individual attribute and object representations are considered to be the output of the encoder when each of these tokens are fed {\it separately} into the encoder, while in the previous experiment, we only analyzed the embedding of the attribute-object composition. We prepared a dataset containing embeddings for object-attributes combinations, as well as separate embeddings for the object and attributes  individually, in both the text and images. We then trained a single-layer network to predict the component embeddings (attribute and object) from the combined embedding, using the linear activation function. We know that for such decomposition to be possible, a sufficient condition is that attributes and objects representations be orthogonal. Let $z$ be the representation of the combination, and further assume that $z = x + y$, with $x$ and $y$ be representations of the attributes and objects, and $x \perp y$. Let $X$ and $Y$ be the orthogonal basis for $x$ and $y$, respectively. Now, $X X^\top z = W_x z = X X^\top x = x$, and $Y Y^\top z = W_y z = Y Y^\top y = y$. Therefore, under such conditions, one could train a linear classifier to uniquely determine $x$ and $y$ from the input $z$. Note that however, if the subspaces that $x$ and $y$ live in are not orthogonal, one could not uniquely determine $x$ and $y$ from $z$. 

For the text representations, we used the embeddings from the CLIP text encoder. For the image representations, we generated images representing each attribute, object, and their combinations using text-to-image models. We then calculated the average embeddings for each attribute, object, and their combination from the CLIP image encoder.

Finally, we determined the test set loss for the single-layer network on both the text and image embeddings. TThe train and test splits contain non-overlapping objects and attributes to ensure compositionality. Notably, the test split consists of combinations of attributes and objects that the model did not encounter during the training phase, thereby assessing the model's ability to generalize to novel compositions. As shown in Figure \ref{fig:orth}.b, models with higher ImageNet-AO accuracy tend to exhibit lower reconstruction loss, indicating better separation of components information in the combined embeddings. Additionally, the loss was lower for the text encoder compared to the image encoder.

We conducted an additional experiment using a two-layer network architecture. In this setup, the weights of the second layer were shared and transposed from the first layer. The results of this experiment, depicted in Figure \ref{fig:orth}.c, closely mirror those of the single-layer network, with higher ImageNet-AO accuracy correlating with lower reconstruction loss. 

These experiments provide insights into the compositional capabilities of CLIP's text and image representations and their potential impact on downstream performance.

\subsection{track disentanglement of embeddings in both
modalities throughout training}

\begin{figure}
  \centering
  \includegraphics[width=0.6\textwidth]{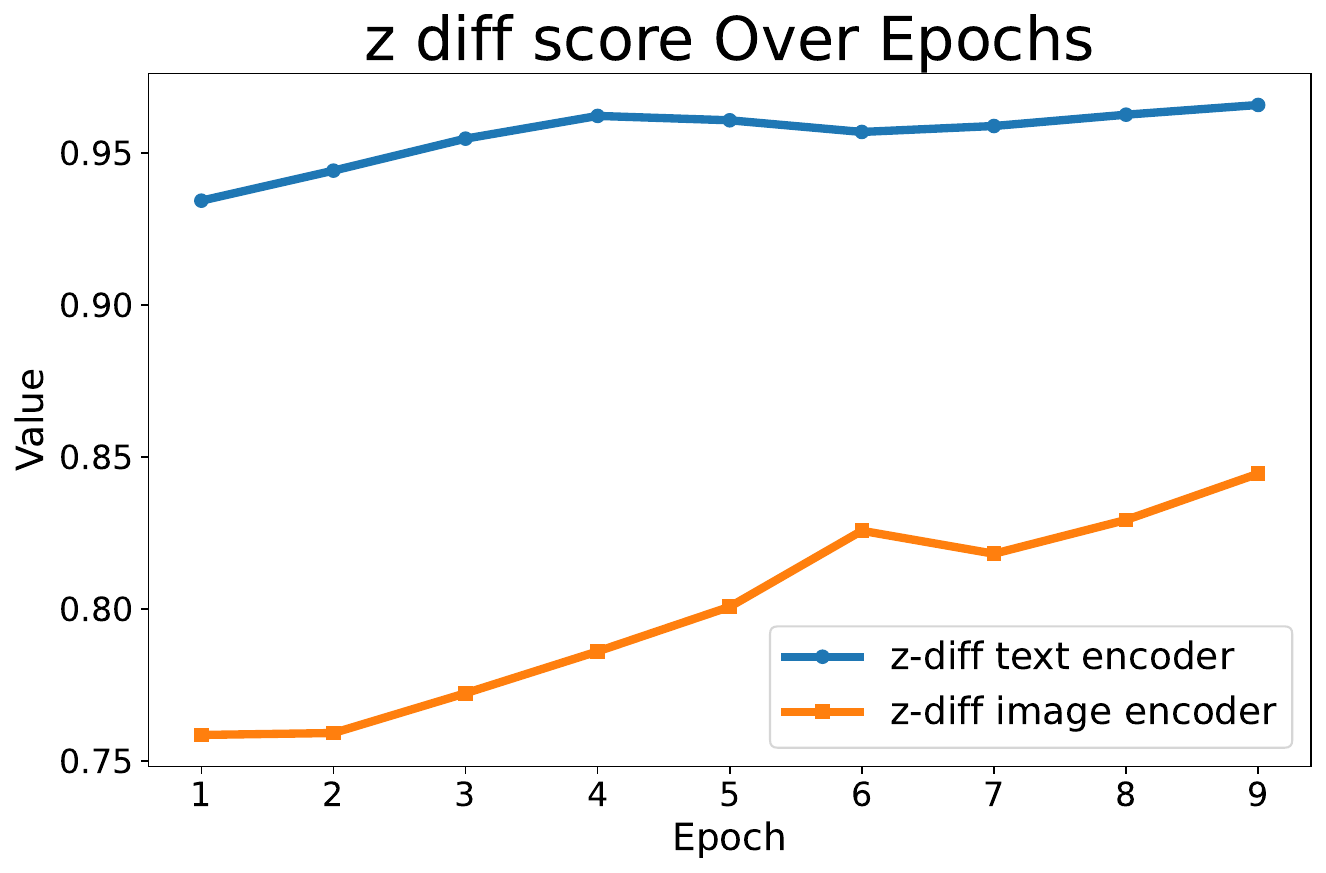} 

  \caption{Fig. R: Z-diff scores for CLIP's encoders trained from scratch.
}
  
  \label{fig:R}
\end{figure}

We trained a CLIP model from scratch on the CC3M dataset, tracking disentanglement (z-diff) for both text and image embeddings at each epoch. As shown in \ref{fig:R}, text embeddings exhibited strong disentanglement early in training, suggesting they drive the disentanglement of image embeddings.

\subsection{ImageNet-AO v2}
With the emergence of GPT-4V and the availability of a powerful model for Visual Question Answering (VQA) tasks, we refined our dataset once again using this model to achieve a higher quality dataset. In this process, for each image related to an attribute-object pair, we presented the image to the GPT-4V model and asked the following four questions:

\begin{enumerate}
    \item Is object X present in this image?
    \item Is attribute Y visible in this image?
    \item Is the combination {Y,X} present in this image?
    \item Is the {Y,X} present in the image rare?
\end{enumerate}

In this operation, we only retained images that received positive answers to all four questions. As a result, we obtained a high-quality subset of our original dataset, filtered by a powerful model like GPT-4V. This subset is available and viewable alongside the original image collection.

Additionally, you can observe the performance results of some models on this dataset in the table \ref{tab:results_part1gpt}.

{\footnotesize 
\begin{table}[t]
\centering
\caption{Performance comparison on ImageNet-AO-v2 (Part 1).}
\label{tab:results_part1gpt}
\adjustbox{max width=\textwidth}{ 
\begin{tabular}{llr}
\toprule
\textbf{Name} & \textbf{Pretrained} & \textbf{Imagenet-AO} \\
\midrule
EVA02-E-14-plus & laion2b\_s9b\_b144k & 87.06 \\
ViT-bigG-14 & laion2b\_s39b\_b160k & 86.05 \\
EVA02-E-14 & laion2b\_s4b\_b115k & 85.02 \\
ViT-H-14-378-quickgelu & dfn5b & 83.66 \\
ViT-H-14 & laion2b\_s32b\_b79k & 83.47 \\
ViT-L-14-CLIPA & datacomp1b & 82.23 \\
convnext\_xxlarge & laion2b\_s34b\_b82k\_augreg\_soup & 84.02 \\
convnext\_xxlarge & laion2b\_s34b\_b82k\_augreg\_rewind & 84.07 \\
convnext\_xxlarge & laion2b\_s34b\_b82k\_augreg & 83.75 \\
xlm-roberta-large-ViT-H-14 & laion\_laion5b\_s13b\_b90k & 78.05 \\
convnext\_xxlarge & laion2b\_s34b\_b82k\_augreg & 83.75 \\
convnext\_large\_d\_320 & laion2b\_s29b\_b131k\_ft\_soup & 83.20 \\
ViT-H-14-CLIPA-336 & laion2b & 83.51 \\
convnext\_large\_d\_320 & laion2b\_s29b\_b131k\_ft & 82.56 \\
ViT-g-14 & laion2b\_s12b\_b42k & 83.05 \\
ViT-H-14-378-quickgelu & dfn5b & 83.18 \\
ViT-bigG-14-CLIPA & datacomp1b & 82.94 \\
ViT-H-14-CLIPA-336 & datacomp1b & 82.16 \\
ViT-H-14-CLIPA-336 & datacomp1b & 83.51 \\
ViT-SO400M-14-SigLIP-384 & webli & 82.15 \\
ViT-B-16-SigLIP-384 & webli & 81.69 \\
ViT-B-16-SigLIP-256 & webli & 81.55 \\
convnext\_base\_w\_320 & laion\_aesthetic\_s13b\_b82k\_augreg & 82.10 \\
ViT-H-14 & laion2b\_s32b\_b79k & 83.47 \\
ViT-B-16-SigLIP-512 & webli & 81.68 \\
coca\_ViT-L-14 & laion2b\_s13b\_b90k & 80.98 \\
ViT-L-14-CLIPA-336 & datacomp1b & 81.75 \\
ViT-L-14 & commonpool\_xl\_clip\_s13b\_b90k & 82.04 \\
convnext\_base\_w & laion2b\_s13b\_b82k\_augreg & 80.91 \\
convnext\_base\_w\_320 & laion\_aesthetic\_s13b\_b82k & 81.49 \\
ViT-H-14-quickgelu & metaclip\_fullcc & 81.17 \\
ViT-L-14-quickgelu & metaclip\_fullcc & 80.81 \\
convnext\_base\_w & laion2b\_s13b\_b82k & 80.91 \\
ViT-B-16-SigLIP-i18n-256 & webli & 81.39 \\
ViT-L-14-quickgelu & dfn2b & 81.70 \\
ViT-L-14 & datacomp\_xl\_s13b\_b90k & 81.87 \\
ViT-B-16-SigLIP & webli & 80.07 \\
coca\_ViT-L-14 & mscoco\_finetuned\_laion2b\_s13b\_b90k & 80.99 \\
convnext\_base\_w & laion\_aesthetic\_s13b\_b82k & 80.91 \\
ViT-L-16-SigLIP-256 & webli & 81.58 \\
\bottomrule
\end{tabular}
}
\end{table}
}

\begin{table}[htbp]
\centering
\caption{Performance comparison on ImageNet-AO-v2 (Part 2).}
\label{tab:results_part2}
\adjustbox{max width=\textwidth}{ 
\begin{tabular}{llr}
\toprule
\textbf{Name} & \textbf{Pretrained} & \textbf{Imagenet-AO} \\
\midrule
ViT-SO400M-14-SigLIP & webli & 81.69 \\
EVA01-g-14 & laion400m\_s11b\_b41k & 80.13 \\
EVA02-L-14-336 & merged2b\_s6b\_b61k & 79.76 \\
ViT-B-16 & datacomp\_xl\_s13b\_b90k & 80.35 \\
EVA02-L-14 & merged2b\_s4b\_b131k & 79.52 \\
ViT-L-16-SigLIP-384 & webli & 81.67 \\
ViT-B-16 & laion2b\_s34b\_b88k & 78.90 \\
ViT-L-14 & commonpool\_xl\_laion\_s13b\_b90k & 80.04 \\
ViT-L-14 & laion400m\_e32 & 79.51 \\
ViT-B-16-quickgelu & metaclip\_fullcc & 81.18 \\
ViT-L-14 & laion400m\_e31 & 79.51 \\
ViT-B-16 & dfn2b & 80.08 \\
ViT-B-32 & laion2b\_s34b\_b79k & 78.49 \\
ViT-B-32 & laion2b\_e16 & 78.51 \\
ViT-B-32-256 & datacomp\_s34b\_b86k & 79.27 \\
roberta-ViT-B-32 & laion2b\_s12b\_b32k & 77.85 \\
xlm-roberta-base-ViT-B-32 & laion5b\_s13b\_b90k & 78.05 \\
ViT-B-32 & datacomp\_xl\_s13b\_b90k & 78.15 \\
ViT-B-16-plus-240 & laion400m\_e31 & 78.97 \\
ViT-B-16-plus-240 & laion400m\_e32 & 78.97 \\
EVA02-B-16 & merged2b\_s8b\_b131k & 77.13 \\
coca\_ViT-B-32 & laion2b\_s13b\_b90k & 76.65 \\
ViT-B-16 & laion400m\_e31 & 77.61 \\
ViT-B-16 & laion400m\_e32 & 77.61 \\
ViT-L-14-quickgelu & metaclip\_400m & 77.82 \\
ViT-B-32-quickgelu & metaclip\_fullcc & 75.28 \\
ViT-B-16-quickgelu & metaclip\_400m & 76.61 \\
ViT-L-14 & commonpool\_xl\_s13b\_b90k & 75.16 \\
ViT-B-16 & datacomp\_l\_s1b\_b8k & 75.35 \\
convnext\_base & laion400m\_s13b\_b51k & 74.71 \\
ViT-B-32-quickgelu & laion400m\_e31 & 75.25 \\
ViT-B-32-quickgelu & laion400m\_e32 & 75.12 \\
ViT-B-32 & laion400m\_e31 & 74.71 \\
ViT-B-32 & laion400m\_e32 & 74.71 \\
ViT-B-16 & commonpool\_l\_clip\_s1b\_b8k & 73.64 \\
ViT-L-14-336 & openai & 72.86 \\
ViT-B-16 & commonpool\_l\_laion\_s1b\_b8k & 71.55 \\
ViT-L-14 & openai & 71.57 \\
ViT-B-16 & commonpool\_l\_text\_s1b\_b8k & 71.08 \\
ViT-B-16 & commonpool\_l\_image\_s1b\_b8k & 70.35 \\
ViT-B-16 & openai & 66.66 \\
RN50x64 & openai & 70.74 \\
RN50x16 & openai & 69.08 \\
ViT-B-16 & commonpool\_l\_basic\_s1b\_b8k & 71.08 \\
ViT-B-32-quickgelu & openai & 66.66 \\
ViT-B-32 & openai & 66.66 \\
RN50x4 & openai & 66.44 \\
\bottomrule
\end{tabular}
}
\end{table}

\end{document}